\documentclass{article}
\usepackage{geometry}
\usepackage[utf8]{inputenc}
\usepackage{subcaption}

\usepackage{cite}

\usepackage{nameref,hyperref}

\usepackage[right]{lineno}

\usepackage{microtype}
\DisableLigatures[f]{encoding = *, family = * }

\raggedright
\usepackage{changepage}

\usepackage[aboveskip=1pt,labelfont=bf,labelsep=period,singlelinecheck=off]{caption}

\makeatletter
\renewcommand{\@biblabel}[1]{\quad#1.}
\makeatother

\usepackage{lastpage,fancyhdr,graphicx}
\usepackage{epstopdf}
\fancyheadoffset[L]{2.25in}
\fancyfootoffset[L]{2.25in}

\usepackage{color}

\definecolor{Gray}{gray}{.25}

\usepackage{graphicx}

\usepackage{sidecap}

\usepackage{wrapfig}
\usepackage[pscoord]{eso-pic}
\usepackage[fulladjust]{marginnote}
\reversemarginpar

\begin{document}
\vspace*{0.35in}

\begin{flushleft}
{\Large
\textbf\newline{Can you even tell left from right? Presenting a new challenge for VQA (Under revision in pattern recognition letters)}
}
\newline
\\
Sai Raam Venkataraman\textsuperscript{1*},
Rishi Rao\textsuperscript{1},
S. Balasubramanian\textsuperscript{1},
Chandra Sekhar Vorugunti\textsuperscript{2},
R. Raghunatha Sarma\textsuperscript{1},
\\
\bigskip
\bf{1} Department of Mathematics and Computer Science, Sri Sathya Sai Institute of Higher Learning
\\
\bf{2} Samsung Research, India
\\
\bigskip
* vsairaam@sssihl.edu.in

\end{flushleft}

\section*{Abstract}
Visual Question Answering (VQA) needs a means of evaluating the strengths and weaknesses of models. One aspect of such an evaluation is the
evaluation of \textit{compositional generalisation}, or the ability of a model to answer well on scenes whose scene-setups are different from the training set. Therefore, for this purpose, we need datasets whose train and test sets differ significantly in composition. In this work, we present several quantitative measures of compositional separation and find that popular datasets for VQA are not good evaluators. To solve this, we present Uncommon Objects in Unseen Configurations (UOUC), a synthetic dataset for VQA. UOUC is at once fairly complex while also being well-separated, compositionally. The object-class of UOUC consists of 380 clasess taken from 528 characters from the Dungeons and Dragons game. The train set of UOUC consists of 200,000 scenes; whereas the test set consists of 30,000 scenes. In order to study compositional generalisation, simple reasoning and memorisation, each scene of UOUC is annotated with up to 10 novel questions. These deal with spatial relationships, hypothetical changes to scenes, counting, comparison, memorisation and memory-based reasoning. In total, UOUC presents over 2 million questions. UOUC also finds itself as a strong challenge to well-performing models for VQA. Our evaluation of recent models for VQA shows poor compositional generalisation, and comparatively lower ability towards simple reasoning. These results suggest that UOUC could lead to advances in research by being a strong benchmark for VQA.

\section{Introduction}
The field of Visual Question Answering (VQA) deals with the development of machine learning models that can understand visual scenes, and answer questions about them. These questions can deal with the properties of objects present in the scenes, or with the relations among the objects. Evaluating models involves measuring their ability to answer questions correctly. Two aspects for this are: a model's ability to handle complex scenes, and its ability to generalise well to scenes whose compositions differ from the scenes it was trained on. These we refer to as \textit{expressivity} and \textit{compositionality}.

Why are these two important? Expressivity is important, as useful data is often complex. Compositionality is important as compositional models learn concepts independent of the objects the concepts are associated with on the train set. To explain this, consider that the concept of 'front of' is independent of objects that are in front other objects. Thus, this relationship is understood across a variety of objects. This is likewise for many properties, such as colour or size. Therefore, learning of such concepts can be measured by evaluating models on data with new objects that showcase same concepts.

Thus, a dataset for this two-fold evaluation must be \textit{complex} and exhibit \textit{compositional separation} between the train and test scenes. The first property necessitates at least a fair diversity in object-complexity. The second is so that the train and test scenes have a minimal overlap in composition. This is so that the learning of concepts can be evaluated outside of instances seen in training. 

\begin{figure}
    \centering
    \begin{subfigure}[t]{0.15\textwidth}
    \includegraphics[width=\textwidth]{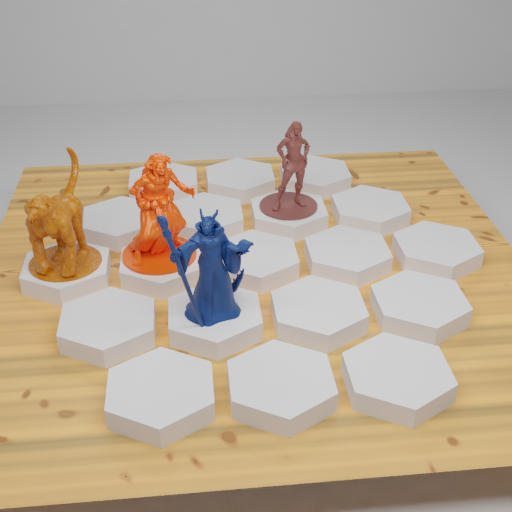}
    \caption{}
    \label{subfig:example_test_scene}
    \end{subfigure}
    \begin{subfigure}[t]{0.23\textwidth}
    \includegraphics[width=\textwidth]{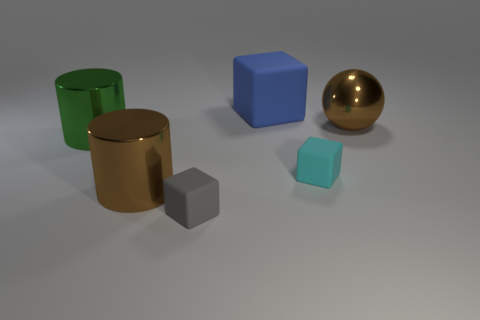}
    \caption{}
    \label{subfig:clevr_example_scene}
    \end{subfigure}
    \begin{subfigure}[t]{0.12\textwidth}
    \includegraphics[width=\textwidth]{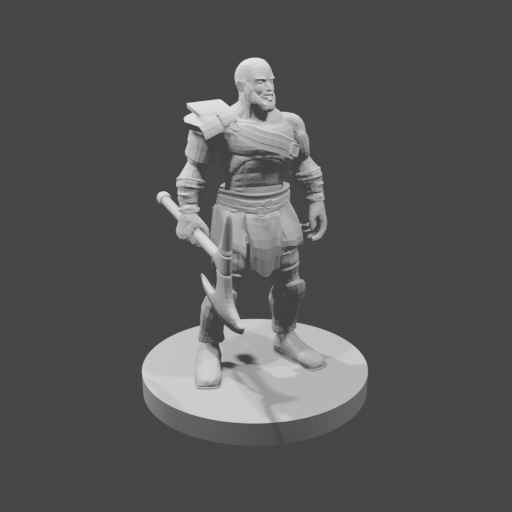}
    \caption{}
    \label{subfig:sample_first_object}
    \end{subfigure}
    \begin{subfigure}[t]{0.12\textwidth}
    \includegraphics[width=\textwidth]{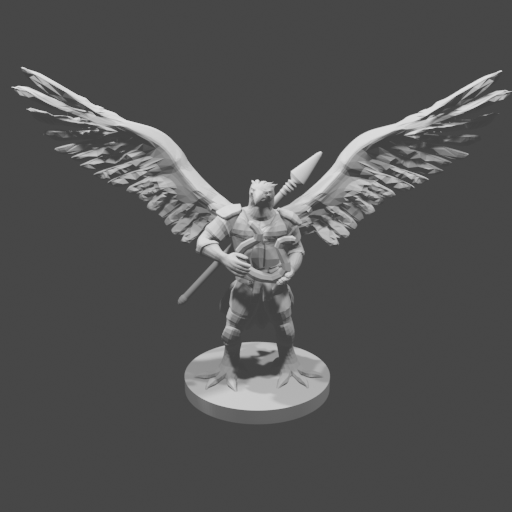}
    \caption{}
    \label{subfig:example_second_object}
    \end{subfigure}
    \begin{subfigure}[t]{0.12\textwidth}
    \includegraphics[width=\textwidth]{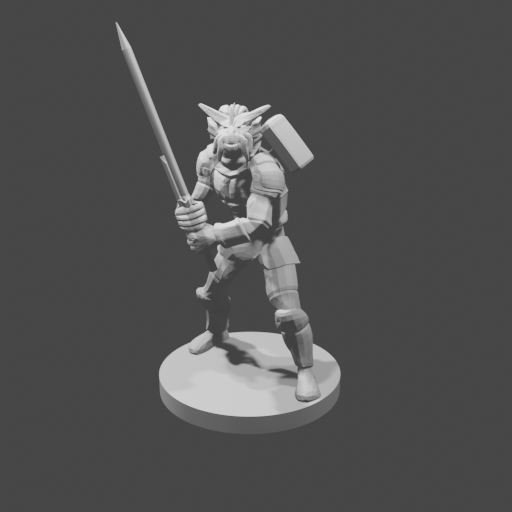}
    \caption{}
    \label{subfig:example_third_object}
    \end{subfigure}
    \caption{Fig. \ref{subfig:example_test_scene} is an example scene in the test set of UOUC. Every pair of objects in it are never seen in the train set - this makes for a strong evaluation of relationships such as spatial ones. Fig. \ref{subfig:clevr_example_scene} is an example scene in CLEVR. Note that the complexity of objects is low, due to them being simple geometric objects of only three classes. In contrast, consider the objects in Fig. \ref{subfig:example_test_scene}, or in Fig. \ref{subfig:sample_first_object}, \ref{subfig:example_second_object} and \ref{subfig:example_third_object}. These possess greater complexity. This, combined with compositional separation, makes for a challenge to models.}
\end{figure}

Datasets that are based on natural scenes are significantly complex. Examples of these are VQA-v1 \cite{VQA}, VQA-v2 \cite{balanced_vqa_v2}, Visual Genome \cite{krishna2017visual} and GQA \cite{Hudson_2019_CVPR}. They, however, do not have a mechanism for compositional separation. Moreover, the presence of natural biases in the co-occurrence structures of objects could lead to the presence of compositional similarities in the train and test datasets.

On the contrary, a dataset such as CLEVR \cite{johnson2017clevr} offers a structured generation of scenes, where certain aspects of the train scenes differ from the test scenes. Specifically, the CoGenT dataset of CLEVR presents a colouring-based compositional separation, where certain objects are coloured differently in the train and test sets. However, CLEVR suffers from the fact that its object-set is small and simple - consisting of cylinders, cubes and spheres. Each of these objects have only three other properties, namely colour, material and size. Thus, despite the generated compositional difference of colouration, train and test sets of CoGenT have several similarities. Another weakness is that due to the simplicity of the object-classes, models may easily learn to recognise objects and thus exhibit high performance.

A major source of this compositional similarity in the train and test sets of not only CLEVR, but also other VQA datasets, is the presence of common co-occurrences of object-pairs. In other words, several object-pairs occur together in scenes of both train and test data. The presence of this common information can make answering questions easier for models, while obscuring the real extent of compositional generalisation. 

As a means of solving this problem, we present Uncommon Objects in Unseen Configurations (UOUC). UOUC is significantly more complex than CLEVR. As a rough comparison, the object-class of UOUC is 380 in number, and comes from 528 characters of the Dungeons and Dragons game. Unlike the simple objects of CLEVR as seen in Figure \ref{subfig:clevr_example_scene}, these exhibit significant complexity as can be seen in the example scene in Figure \ref{subfig:example_test_scene} and the example objects in Figures \ref{subfig:sample_first_object}, \ref{subfig:example_second_object} and \ref{subfig:example_third_object}. More scenes are present in the supplementary material. UOUC borrows the idea from CLEVR that synthetic generation is a means of designing compositionally-separated train and test sets.

Unlike CLEVR, however, UOUC does not use a simple colouration-based condition. Instead, UOUC is designed so that no pair of objects co-occur in both the train and test sets. Thus, the scenes of the test set are compositionally different in a more complex manner.

In order to use this separation condition effectively, each scene of UOUC is annotated with up to 10 questions. The first four of these deal with the learning of spatial relationships between objects, namely front, back, left, and right. Since no objects co-occur commonly, a model must learn to understand the notion of a spatial relationship independent of the instances it has seen during training. The next two questions deal with simple reasoning such as counting and comparison. The next question deals with pure perception, and queries the presence of an object in a scene. The last three questions deal with memorisation and memory-based reasoning. Memorisation related to non-perceptual attributes is an under-explored area of VQA. Specifically, we wish to make an initial direction in understanding the complexity of memory-based reasoning, with respect to perception-based reasoning.

UOUC thus consists of 200,000 questions in the training set, and 30,000 questions in the test set, with over 2 million questions for all the scenes.

Our experiments using recent models show that compositional generalisation is hard, whereas simple reasoning is relatively easy. Memory-based reasoning is still easier for the best-performing models. In general, UOUC offers a challenge for VQA in terms of compositional generalisation, that could lead to advances in the field.

The organisation of the rest of the paper is as follows. Section 2 presents related work. Section 3 presents a quantitative comparison of UOUC with other datasets. Section 4 presents a description of UOUC in terms of its construction, object-classes and their attributes, and associated questions. Section 5 presents a description of experiments done using recent models on UOUC. Section 6 presents a discussion of these results. We then present a discussion of possible future work and extensions of UOUC.

\section{Related Work}\label{rela}
There is much work in the field of VQA involving contributions in terms of data sets and models. Several models have been proposed and evaluated on various data sets. MUTAN \cite{ben2017mutan} is a model that combines visual and question features in an efficient manner, achieving good performance on VQA-v1. Likewise, \cite{kazemi2017show} presented a model that combines features from CNNs, attention, and a sequence-model to again achieve good performance on VQA-v1 and VQA-v2. More recently, LCGN and MAC \cite{hudson2018compositional} were proposed. LCGN \cite{hu2019language} is a language-conditioned graph network that uses iterative message-passing for VQA. MAC is a compositional model for VQA that uses memory and control, integrated in a single unit. LCGN and MAC achieve good accuracies on CLEVR and GQA. 


Data sets for VQA that many models for VQA use are VQA-v1, VQA-v2, GQA, Visual Genome, and CLEVR. VQA-v1 was one of the early data sets for VQA. It uses images from the COCO data set \cite{chen2015microsoft}, and provides annotations in the form of questions. VQA-v2 is a balanced version of this data set that associates each question in VQA-v1 with two images that have a different answer. Visual Genome is a highly-annotated data set of real images for VQA, with a huge object-class. GQA is a data set of real scenes that annotates images with questions that involve multiple steps of reasoning. The answer distribution of the questions has been balanced. Moreover, GQA also introduces metrics for evaluating models that move beyond accuracy. CLEVR is a generated data set for VQA that uses questions that are compositional in nature, and can use multiple steps of reasoning. CLEVR offers a data set termed CoGenT that uses a separation condition, in terms of object-properties, in its train and test set. Compositional models that understand properties independent of objects would be able to answer questions that relate to the properties of this separation condition.



\section{How does UOUC compare to popular datasets}
In this section, we present a comparison of UOUC with several popular datasets for VQA. Our comparison, like our motivation for UOUC, is on the two fronts of complexity and compositional separation. 

\subsection{Comparing the richness and complexity}
We aim to provide a measure of complexity, or richness, by stating the number of questions, objects, and scenes in the datasets. Table \ref{tab:numimq} provides this information. As can be seen, UOUC has an object set much larger than CLEVR-CoGenT. Moreover, and as mentioned before, based on a comparison between objects in Figure \ref{subfig:example_test_scene} and Figure \ref{subfig:clevr_example_scene}, we can safely say that UOUC has more complex scenes. UOUC is also quite comparable in terms of the number of scenes, the size of the object-class and number of questions to datasets such as VQA-v1 and VQA-v2. This establishes UOUC as a perceptually harder synthetic dataset for VQA than CLEVR.

\subsection{Comparing compositional separation}
As stated before, two measures of compositional separation between the train and the test data are the number of common co-occurring pairs and relations among them. We use this idea to present four metrics that, when close to zero, indicate a high degree of compositional separation.

\subsubsection{Common co-occurrences}
The first of these, termed \textit{AvgCoPair}, is the average number of co-occurring pairs of objects in a test scene, that have co-occurred in a train scene. Complementing this is a second score, termed \textit{AvgCoPairOcc}, which gives the average number of times a commonly co-occurring pair is found in the train dataset.

\begin{table}[]
    \centering
    \begin{tabular}{|c|c|c|c|}
         \hline
         Name & Images & Questions & Classes\\
         \hline
         VQA-v1 (Real)& 205K& 614K &80\\
         \hline
         VQA-v2 (Real)& 205K& 1.1M & 80\\
         \hline
         Visual Genome& 108K& 1.7M & 33,877 \\
         \hline
         GQA & 113K & 22M & 1703 \\
         \hline
         CLEVR-CoGenT & 130K &  1.3M & 3 \\
         \hline
         \textbf{UOUC} & \textbf{230K} & \textbf{2M} & \textbf{380} \\
         \hline
    \end{tabular}
    \caption{The number of images, objects,  questions for some data sets (approximate).}
    \label{tab:numimq}
\end{table}

\begin{table}[]
    \centering
    \begin{tabular}{|c|c|c|c|c|}
         \hline
         Data set & Visual Genome& GQA & \textbf{UOUC} & CoGenT\\
         \hline
         AvgCoPair & 269.93 & 155.95 & \textbf{0.00} & 20.89 \\
         \hline
         AvgCoPairOcc & 2,270.14 & 4,023.83 & \textbf{0.00} & 318K \\
         \hline
         AvgCoRel & 6.20 & 45.79 & \textbf{0.00} & - \\
         \hline
         \hline
    \end{tabular}
    \caption{Measure of compositional separation between the train and test sets. Lower indicates better separation. Note that UOUC is the best separated among all the datasets, making it an ideal dataset for evaluating compositional generalisation.}
    \label{tab:relsc}
\end{table}

The first measure gives an extent of common information, per test image, present between the train and test set. The second measure gives the extent of the presence of this common information in the train set. For example, if two objects - a child and a cake, were to be present in the same scene for some images in both the train the and test datasets, there is some compositional overlap. This compositional overlap can lead to models obtaining higher performance because they have seen these together in the train dataset. Moreover, as mentioned before, this also leads to a difficulty in measuring compositional generalisation for relationships that may have existed between these. If multiple such instances of such co-occurrences are found in the train dataset, then it becomes easier for a model to memorise this information and then answer questions on the test dataset for the common pairs.

\subsubsection{Common relationships}
The next two measures, termed AvgCoRel and AvgCoRelOcc, extend this concept to relationships between objects instead of only co-occurrences. The presence of the same relationships between the same objects in both the train and test set is a stronger case of compositional overlap. Further a larger number of occurrences of any such overlap in the train dataset makes their learning easier for models, and thus impacts a proper evaluation.

Using the same example of a child and a cake, if both the train and the test data have common instances of the relation 'eating' between them, there is a strong compositional overlap. If multiple instances of a child eating cake exist in the train dataset, then a model can find it easy to use this in answering questions on the test dataset.

Table \ref{tab:relsc} gives these four metrics for Visual Genome, GQA, UOUC, and CLEVR. We use the mean of 5 random 70-30 train-test splits for Visual Genome to obtain these results, in the manner of \cite{krishna2017visual}. We use the validation set as the test set for GQA, as required information about the test set is not provided publicly. As CLEVR uses no annotated relationships between objects, we compute only AvgCoPair and AvgCoPairOcc. Lastly, as VQA-v1 and VQA-v2 do not present required information for the computation of these four scores, we do not present them.

First, we see that Visual Genome, GQA, and CLEVR have non-zero scores for the computed measures. Thus, there is a high level of compositional overlap between the train and test splits. UOUC has all the scores zero, as no common co-occurring pairs or relationships exist. This is by design. Thus, UOUC offers a stronger evaluation of compositional generalisation for models.

\subsection{Comparison by performance}
A final comparison of datasets is based on the level of challenge it offers to recent models. Such a comparison can be based on the accuracy of answering questions. Based on Table \ref{tab:accuracies}, we see that models that achieve decent performance on VQA-v2 and GQA, and more importantly high performance on CLEVR, perform poorly on UOUC, especially on compositional generalisation. This allows us to state that UOUC can lead to further advances for compositional generalisation in VQA.

\section{Details of UOUC}
UOUC, like CLEVR, is synthetically generated so as to allow for a compositional separation of the train and test sets. We outline the process of construction. 

\textbf{Downloading and pre-processing object models:} All the objects were downloaded from \url{https://www.prusaprinters.org/social/39782-mz4250/prints} as .STL files. They were pro-processed to make the scales uniform, and the the 3d models of the objects face the viewer. In total, 528 models were used to create 380 classes of objects. The objects are used with the permission of the creator and are licensed under a Creative Commons License 4.0 (Non-Commercial) license.

\textbf{Categorisation and grouping of objects:} 
The 380 classes were further categorised into 5 categories. These categories are presented in bold in Table \ref{tab:categories}. The categorisation was done manually. After categorisation, the objects are again grouped randomly into 10 groups. The process is such that each group has roughly similar numbers of objects of each category. This grouping does not introduce any new properties to objects, but is extremely important to the compositional separation of the train and test sets.

\textbf{Construction of the base scene:} Since the scenes are to be generated using the 3d models of the objects, we use the Blender \cite{blender} software to first create 3d scenes which are then rendered to 2d scenes. A requisite for this is a 3d base scene. Figure \ref{fig:base_scene} shows the scene we used. The scene is made so that the objects are placed in the 19 hexagonal structures.

\textbf{Construction of the scene-structures:} Before generating the scenes of the train and test sets, we first generate text-files that store their structures. The structure of a scene consists of the positions of each of the objects in the scene, the colour of the objects, their rotation about the z-axis, and the camera-rotation. For the train set set, objects are rotated with an angle chosen randomly between -30 and 30 degrees; the camera is rotated randomly with an angle chosen between -20 and 20 degrees. For the test set, these extents are -45 and 45 degrees for the objects, and -30 and 30 degrees for the camera. These are stored in the text-files, which are then used for the generation of the scenes and the questions and answers. The grouping mentioned earlier is used here. For the train set, scenes are generated so that only objects within a group co-occur, while the test set has that objects only from different groups co-occur. The choice of the objects is otherwise random, with a scene having at least 4 objects, and at most 6. Thus, the train and test sets compositionally-separated. Each group generates 20,000 scenes, and the test set contains 30,000 scenes.

\textbf{Generation of the scenes:} The scenes of UOUC are generated using the scene-structures and the blender software. Based on the scene-structure objects are assigned a colour from one of orange, yellow, green, violet, brown, and light-yellow. The objects are then rotated and positioned according to the scene-structure for that scene. 

\textbf{Generation of the questions and answers:} The text-files find another use in the generation of the questions and answers for the scenes. Each scene of UOUC has up to 10 questions associated with it. The procedure for the generation of questions and answers is based on using predefined templates for each question and filling-in these templates using the logic for that question, and using certain properties of objects. These properties, apart from colour, are mentioned in Table \ref{tab:categories} for each category. The properties are all categorical, and can assume more than 2 values across the objects of a category. The assignment of these properties is based on the role of the object-classes in literature and media. This was done manually. The reader may note that no prior knowledge of Dungeons and Dragons is necessary for using UOUC. Apart from these properties, each object is also assigned a random team, which is one of 'A', 'B', and 'C'. The purpose of this attribute is test the memorisation ability of models.

\begin{figure}
    \centering
    \begin{subfigure}[t]{0.15\textwidth}
    \includegraphics[width=\linewidth]{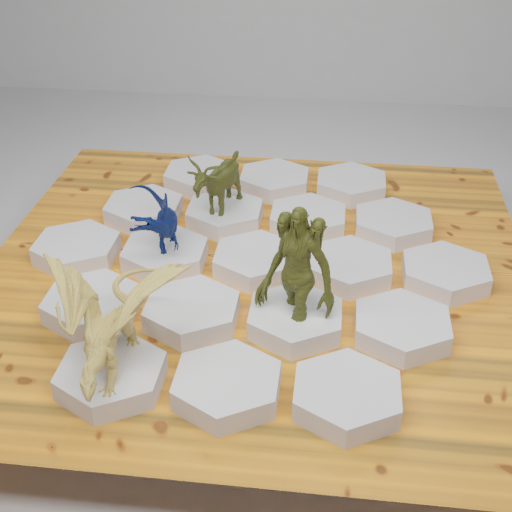}
    \caption{}
    \label{fig:example_of_scene}
    \end{subfigure}
    \begin{subfigure}[t]{0.15\textwidth}
    \includegraphics[width=\linewidth]{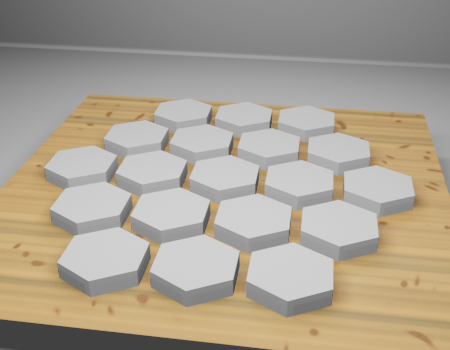}
    \caption{}
    \label{subfig:base_scene}
    \end{subfigure}
    \begin{subfigure}[t]{0.15\textwidth}
    \includegraphics[width=\linewidth]{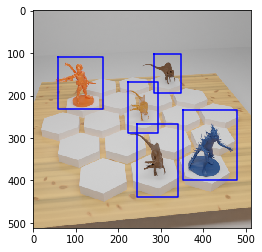}
    \caption{}
    \label{subfig:bounding_box}
    \end{subfigure}
    \caption{Examples of each question-type are given, in order in Figure \ref{fig:example_of_scene}. \textbf{Q1.} Is a green regular-animal front of a blue spinosaurus? \textbf{A1.} no. \textbf{Q2.} Is there a light-yellow nithe dragon left of a blue  spinosaurus that is back of a green goat? \textbf{A2.} no. \textbf{Q3.} Is there a green  goat back of a light-yellow nithe dragon and right of the green  coven-horror? \textbf{A3.} no. \textbf{Q4.} Swapping the position of a green  coven-horror with a green  goat, is a light-yellow nithe dragon front of it? \textbf{A4.} yes. \textbf{Q5.} Are there greater, equal or lesser number of dinosaurs than dragons? \textbf{A5.} equal.
    \textbf{Q6.} Upon removal of blue dinosaur how many blue dinosaur are present? \textbf{A6.} 0. \textbf{Q7.} Is there a orange abeloth dragon in the scene? \textbf{A7.} no. \textbf{Q8.} What is the predation-level of the green goat? \textbf{A8.} 1. \textbf{Q9.} Will goat  attack spinosaurus? \textbf{A9.} no. \textbf{Q10.} Which category does nithe belong to? \textbf{A10.} dragon. Figure \ref{subfig:base_scene} shows the base scene. Figure \ref{subfig:bounding_box} shows a scene with bounding boxes for the objects in the image.}
\end{figure}

We describe each of the questions below. Broadly, they can be categorised into 4 categories: Spatial relationship-based (\textbf{SRB}), perceptual (\textbf{P}), simple reasoning (\textbf{SR}), and memory-based (\textbf{MB}). Examples of each of the questions are given, in order of introduction, in Figure \ref{fig:example_of_scene}.

\begin{table}[h!]
\begin{tabular}{|p{0.2\textwidth}|p{0.2\textwidth}|p{0.2\textwidth}|p{0.2\textwidth}|p{0.2\textwidth}|}
 \hline
 \textbf{Adventurer} & \textbf{Dragon} & \textbf{Animal} & \textbf{Monster} & \textbf{Mythical} \\
 \hline
 Species    & Name & Type            & Type & Type \\
 \hline
 Class      & Pose & Predator-level & Name & Name \\
 \hline
 Gender     &  -  & Prey-level      & -    & Gender \\
 \hline
 Weapon     &  -   & Name            & -    & - \\
 \hline
 Mount      &  -   & Food-habit      & -    & - \\
 \hline
 Attackable &  -   & -               & -    & - \\
 \hline
 \end{tabular}
 \caption{Categories and their attributes.} \label{tab:categories}
\end{table}

\textbf{Q1. Checking for relationships (\textbf{SRB})} This question presents a description of two objects and then asks if a spatial relationship exists between them.

\textbf{Q2. Checking for chains of relationships (\textbf{SRB})} This question provides the description of three objects, and asks if the first satisfies a given spatial relationship with the second, and the second satisfies a second given spatial relationship with the third.

\textbf{Q3. Checking for satisfying relationships of two types (\textbf{SRB})}
This question provides the descriptions of three objects, and asks if the first satisfies a given spatial relation with the second, and also a given second relationship with the third.

\textbf{Q4. Text-only swapping and checking for a relationship (\textbf{SRB})}
This question provides a description of three objects, suggests a swap of position between two, and asks if the third satisfies a given spatial relationship with the first after swapping. This is new kind of question, similar to some suggested in \cite{beckham2020visual}, that changes the scene in text, and asks a question. A model that has learnt to separate an entity and relationships it may be in would find it easier to answer this, than a model that has not.

\textbf{Q5. Comparing based on attributes (\textbf{SR})}
This question gives two descriptions of objects, and asks if the number of objects satisfying the first is lesser than, greater than, or equal to the number of objects satisfying the second.

\textbf{Q6. Count of text-only removal of object (\textbf{SR})} This question suggests a text-only removal of an object, satisfying a description. Then it asks the count of objects satisfying another description. This question is also based on a text-only change of a scene, that necessitates a model understand the concept of removing an object, and counting.

\textbf{Q7. Checking for an object (\textbf{P})}
This question provides a description of an object and asks if that object is there in the scene.

\textbf{Q8. Stating the properties of an object (\textbf{MB})}
This question provides a description of an object and asks a property of that object.

\textbf{Q9. Checking for non-spatial relationships (\textbf{MB})}
Apart from spatial relationships, animals are related to other animals and adventurers by a property-based relationship. Each animal has a predation-level and a prey-level, indicative of some notion of which animal is likely to attack. An adventurer has a property of being attackable, which gives if an animal of sufficient predation-level can attack it. An animal can attack an adventurer if its predation-level is greater than 2 and if the adventurer is attackable. An animal can attack another animal if its predation-level exceeds the prey-level of the other animal. This question gives a memory-based relationship that tests memory-based compositional generalisation.

\textbf{Q10. Stating the category or team of an object (\textbf{MB})}
This question provides a description of an object and asks the category or team of the object. While answering this question can be done by only text, it is useful in teaching a model the properties it needs for memory-based compositional generalisation.

\textbf{A note about memory-based properties and questions:} Why are questions that require text-only memorisation included in the VQA tasks of UOUC? For one, we wish to use these as a sanity-check for the models. Secondly, we wish to study if VQA models can also use visual aspects to improve upon text-only processing. Thirdly, we wish to study if memory-based tasks are harder than perceptual tasks. Note that the main challenge of UOUC is given by the first 4 questions, and then the next 3 questions. The memory-based questions are primarily investigative in nature, and can be used as a sanity-check for models.

Apart from these annotations, 2d and 3d bounding boxes for each object in the scenes of UOUC have been provided. An example is provided Figure \ref{subfig:bounding_box}. The bounding boxes could be used for object-detection features.

\subsection{Certain statistics for UOUC:}
We present the following statistics for UOUC in Figure : the proportion of each category in the object-classes, the proportion of the number of object-instances category-wise, and the proportion of the mean of the instances of each object, seen category-wise.

\begin{figure}
    \centering
    \begin{subfigure}[t]{0.30\textwidth}
    \includegraphics[width=\linewidth]{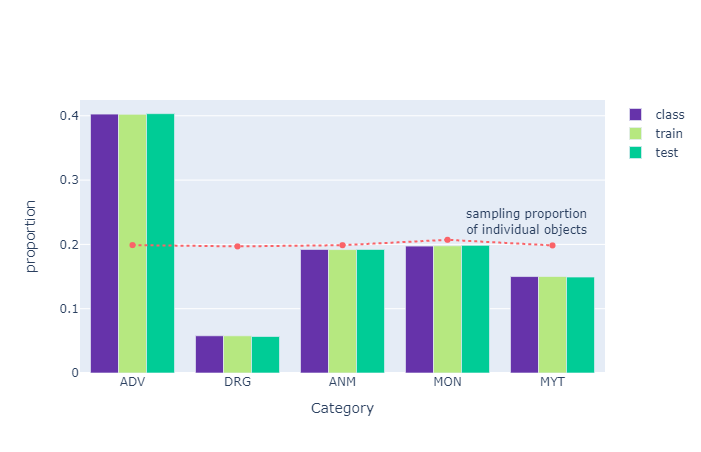}
    \caption{}
    \label{subfig:distribution_objects}
    \end{subfigure}
    \begin{subfigure}[t]{0.30\textwidth}
    \includegraphics[width=\linewidth]{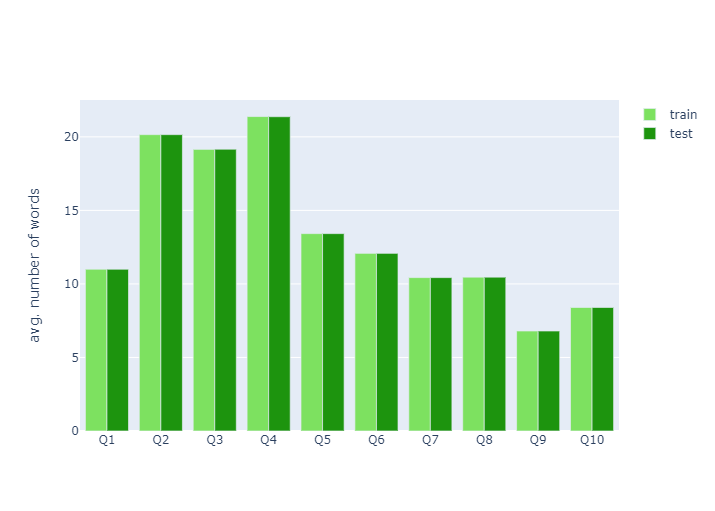}
    \caption{}
    \label{subfig:word_length}
    \end{subfigure}
    \caption{Figure \ref{subfig:distribution_objects} gives the proportion of objects per category (purple) bar, the proportion of object-instances per category for the train set (light-green bar), and the proportion of object-instances per category for the test set (cyan bar). The dotted line shows the distribution of the mean number of instances for each object, seen category-wise. Figure \ref{subfig:word_length} shows the average word-length per question-type, for the train and test set (figures are best viewed in colour).}
\end{figure}

We see that the category 'Adventurers' has the maximum proportion of objects and object-instances. This is because the original data of the 3d models had a large number of such objects. However, due to the random and unbiased sampling of objects in the scenes, each object, regardless of category, can be seen to be almost as equally likely to be present in a scene. Moreover, the distributions of the object-instances and categories are similar for both the train and test sets, ensuring that the models are challenged primarily based on question-answering, rather than by other factors relating to the sampling of objects.

We also plot the average word-length per question-type for both the train and test sets in Figure. We see that the question-lengths are not particularly long, and that they are similar for both the train and test sets. This again ensures that the challenge of answering questions on the test-set of UOUC is based on the logic of the questions, rather than other extraneous factors. Other statistics are given in the supplementary material.

\textbf{Availability of UOUC:} UOUC will be publicly available after the publication of this paper.

\section{Experiments on recent VQA models}
\begin{figure}
    \centering
    \begin{subfigure}[t]{0.30\textwidth}
    \includegraphics[width=\linewidth]{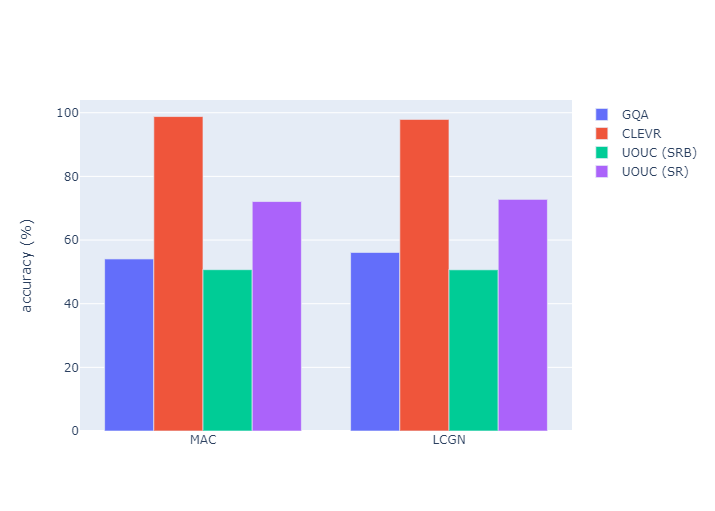}
    \caption{}
    \label{subfig:comparison_mac_lcgn}
    \end{subfigure}
    \begin{subfigure}[t]{0.30\textwidth}
    \includegraphics[width=\linewidth]{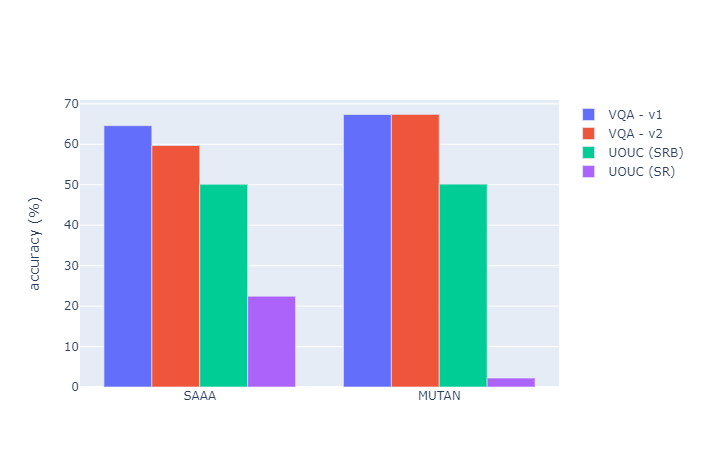}
    \caption{}
    \label{subfig:comparison_saaa_mutan}
    \end{subfigure}
    \caption{Figure Comparison of performances of models on various VQA datasets for VQA and the two main families of questions of UOUC - SRB and SR (figures are best viewed in colour). \ref{subfig:comparison_mac_lcgn} compares the performance of MAC and LCGN on GQA (purple), CLEVR (red), SRB(cyan), and SR (violet). Figure \ref{subfig:comparison_saaa_mutan} compares the performance of SAAA and MUTAN on VQA-v1 (purple), VQA-v2 (red), SRB (cyan), and SR (violet).}
\end{figure}

We trained and tested 4 recent VQA models on UOUC. These are MAC \cite{hudson2018compositional}, LCGN \cite{hu2019language}, SAAA \cite{kazemi2017show}, and MUTAN \cite{ben2017mutan}. MAC and LCGN have achieved high accuracies on CLEVR, and decent performances on GQA. SAAA and MUTAN have achieved decent performances on VQA-v1 and VQA-v2. These results are presented in Figure \ref{subfig:comparison_mac_lcgn} and Figure \ref{subfig:comparison_saaa_mutan}.

In order to use a backbone for the visual features, we trained a ResNet-50 and a ResNet-101 on the problem of detecting the presence of objects in a scene. The precision and recall for these models are: 0.41 and 0.44 for resnet-50, and 0.51 and a recall of 0.32 for resnet-101. Since the ResNet-50 achieves better performance , we use it as the backbone for all the VQA models. This causes a slight deviation from some of the models such as LCGN, which uses a ResNet-101 in the original work.

\begin{figure}
    \centering
    \includegraphics[width=0.40\textwidth]{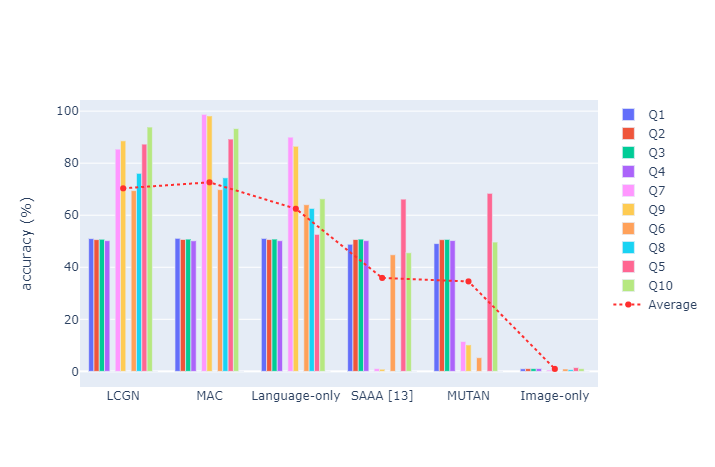}
    \caption{The accuracy per question-type for each model. The average accuracy is given by the dotted line (figure is best viewed in colour).}
    \label{fig:accuracies}
\end{figure}

\begin{table}[]
    \centering
    \begin{tabular}{|c|c|c|c|c|c|c|c|}
         \hline
         Model & \textbf{SRB} & \textbf{SR} & \textbf{P} & \textbf{MB}\\
         \hline
         MAC & 50.69 & 72.10 & 93.30 & 87.69\\
         \hline
         LCGN & 50.65 & 72.78 & 93.88 & 87.10\\
         \hline
         SAAA & 50.13 & 22.45 & 45.57 & 22.68\\
         \hline
         MUTAN & 50.16 & 2.22 & 49.69 & 30.01\\
         \hline
         l-only & 50.69 & 63.32 & 66.30 & 76.38\\
         \hline
         i-only & 1.03 & 0.76 & 1.02 & 0.92\\
         \hline
    \end{tabular}
    \vspace{0.5cm}
    \caption{Accuracies (in\%) for the models on UOUC. Performance on the various types of questions are presented. l-only refers to language only. i-only refers to image-only.}
    \label{tab:accuracies}
\end{table}

We further trained an image-only model and a language-only model for the VQA task. This was done in order to estimate the extent of any biases in the images and questions towards answering the questions. The image-only is a standard CNN-model, with a fully-connected layer for classification. It uses the ResNet-50 as a pre-trained backbone. The language-only model is a transformer model, with a fully-connected layer for classification. A low performance of these models can be inferred as some evidence for lower bias in the dataset. Since their purpose is only to indicate bias, we have not reported results on the other datasets.

Certain training details, like the source of their implementations and the number of epochs, for all these models are given in the supplementary material. The accuracies for the models for each question, along with the average accuracies, is given in Figure \ref{fig:accuracies}. The average accuracy for each family of questions (\textbf{SRB}, \textbf{P}, \textbf{SR}, and \textbf{MB}) for each model is given in Table \ref{tab:accuracies}.

\section{Discussion and analysis of the results}
\subsection{A general analysis of accuracies}
A general analysis of the performance of the models, as seen in Table \ref{tab:accuracies} and
Figure \ref{fig:accuracies}, is that the best-performing models are MAC and LCGN. We also see that SAAA and MUTAN perform poorly in most questions, in contrast to their decent performances on VQA-v1 and VQA-v2, as seen in Figure \ref{subfig:comparison_saaa_mutan}. We hypothesise that the reason for this performance is that SAAA and MUTAN are not able to generalise to the image and question-distribution of the test-set. Also, the image-only model performs poorly, as seen in Figure \ref{fig:accuracies} and Table \ref{tab:accuracies}, indicating low image-bias. We, therefore, do not discuss beyond this in detail for them, focusing on the best-performing models and language-bias. 

For the rest of our discussion, we base our observations on Table \ref{tab:accuracies} and Figure \ref{fig:accuracies}. The comparison between datasets we present follows Figure \ref{subfig:comparison_mac_lcgn}.

\subsection{Analysis of performance per-question}
\textbf{Spatial relationship-based}
Answering questions 1, 2, 3, and 4 requires models to be able to generalise compositionally with respect to spatial relationships. All the models perform almost randomly on these questions. This indicates an inability to learn the spatial relationships in this more complex setting, as well as an inability to generalise compositionally. We believe that these four questions could be a good evaluator of compositionality. This is more so, considering that MAC and LCGN achieve close to 99\% accuracy on CLEVR, and a non-random performance on GQA. We also see that the language-only model perform poorly on these questions, indicating little bias. 

\textbf{Simple reasoning-based}
Answering questions 5 and 6 requires a model to be able to reason based on comparison and counting. We see that MAC and LCGN perform reasonably well at this, achieving over 70\%. However, there is yet a significant scope for improvement. This, again, indicates that UOUC is a challenging dataset for VQA. An interesting point to note is that, while both question 4 and question 6 deal with hypothesised changes to scenes, the best-performing models achieve non-random performance on question 6 alone. This could be seen as evidence that models can even learn to understand complex tasks, such as text-only changes, but compositional generalisation is not easy to demonstrate. The language-only model performs at slightly higher accuracy for these questions, indicating some language-bias. However, this is not significantly high, and the gap between the best-performing models and this model are significant. This shows that the bias is not high enough that the questions can be answered with high accuracy from text alone.

\textbf{Perception}
Question 7 evaluates a model's ability to recognise objects in a scene, given a description. We see that MAC and LCGN perform very well, exceeding 90\% on this question. This leads us to two conclusions. One, that pure perception is comparatively easy for advanced models. Two, that the low performance on the previous questions is not purely due to the complexity of the objects. In other words, the compositional separation between the train and test sets poses a greater challenge than the complex appearances of objects. The language-model achieves around 66\% accuracy, which indicates certain bias in the question-generation. However, again, the gap between the best-performing models and the language-models is quite large, indicating that visual cues are needed to achieve good accuracy.

\textbf{Memory-based}
Questions 8, 9, and 10 can be answered by text alone. However, visual cues can help in the answering of question 9. We see that MAC and LCGN perform well on all these questions, as does the language-only model. Surprisingly, SAAA and MUTAN achieve non-random performance on question 9. We believe further investigation is necessary in understanding this behaviour. We also guess that MAC and LCGN achieve are able to use visual cues along with the text to achieve good performance on question 9. The language-only model performs poorly at memory-based compositional generalisation. Thus, another guess for the higher performance of MAC and LCGN on question 9 could be the presence of 'reasoning-based' architectural design. Again, more investigation is necessary to understand this.

\section{Conclusion}
We introduce a dataset, UOUC, to evaluate compositional generalisation in VQA models. We show that existing datasets do not simultaneously offer complex perceptual tasks and strong evaluations of compositonal generalisation. UOUC fills this need, by presenting a reasonable large and complex object-set, while ensuring compositional separation between the train and test datasets. 
UOUC has 200,000 train scenes and 30,000 test images, with objects drawn from an object set of 380 classes that come from 528 characters of the Dungeons and Dragons game. UOUC is constructed such that no objects co-occur commonly in the train and test sets.

Evaluation of models on UOUC shows that more understanding of compositional generalisation is needed for models, and that UOUC is a strong challenge for the same. We also investigated the use of memorisation questions in VQA. The low performance of certain VQA models suggests their use as a sanity-check. Improvements to UOUC can be done by considering more complex, game-style questions. This would broaden the scope of models that can be evaluated using UOUC. Other improvements can be a harder dataset that uses free positioning. However, currently the easier version is still a hard challenge. We believe that using UOUC as a test for compositional generalisation will allow for more trust in the performance of VQA models.

\section{Supplementary material for the paper}
\section{More scenes of UOUC (in reference to section 1)}
\begin{figure*}
\begin{center}
\begin{subfigure}[t]{0.15\textwidth}
\includegraphics[width=\linewidth]{images/group0scene0.png}
\end{subfigure}
\begin{subfigure}[t]{0.15\textwidth}
\includegraphics[width=\linewidth]{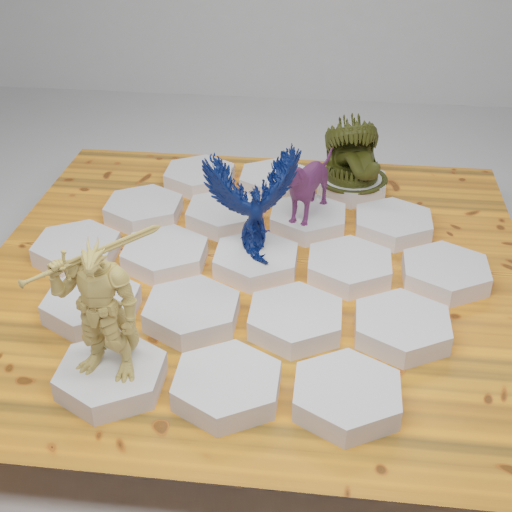}
\end{subfigure}
\begin{subfigure}[t]{0.15\textwidth}
\includegraphics[width=\linewidth]{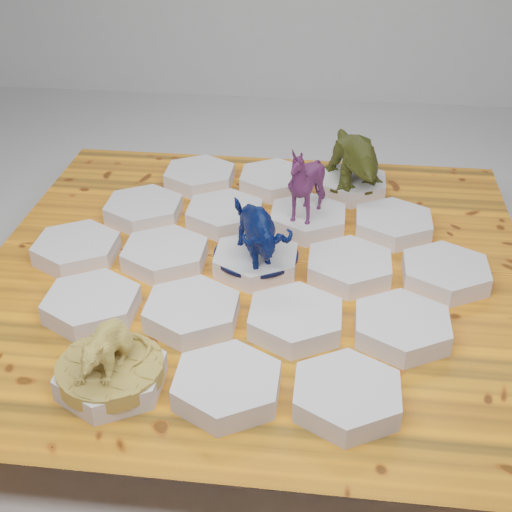}
\end{subfigure}
\begin{subfigure}[t]{0.15\textwidth}
\includegraphics[width=\linewidth]{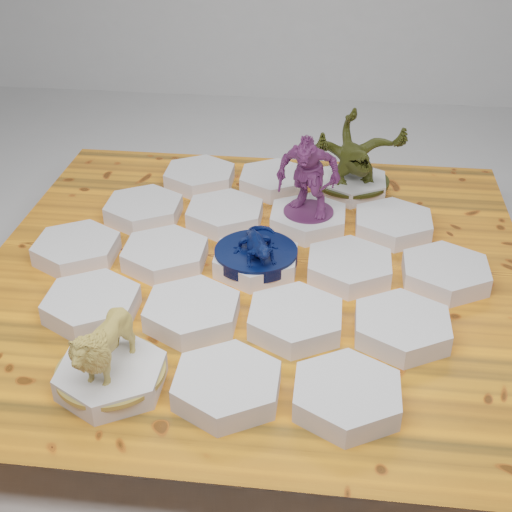}
\end{subfigure}
\begin{subfigure}[t]{0.15\textwidth}
\includegraphics[width=\linewidth]{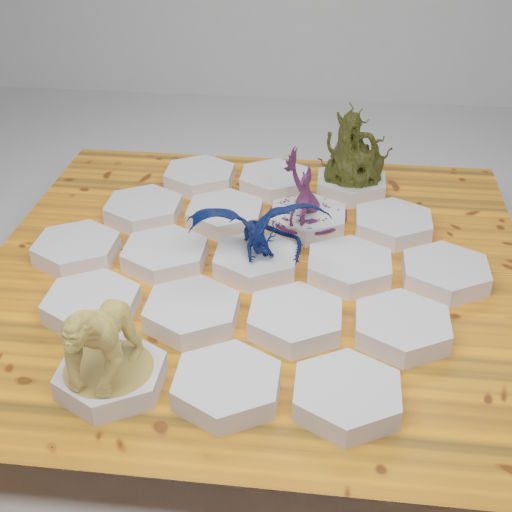}
\end{subfigure}
\begin{subfigure}[t]{0.15\textwidth}
\includegraphics[width=\linewidth]{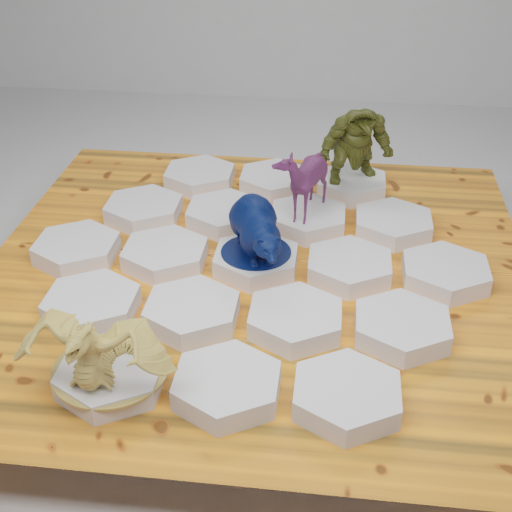}
\end{subfigure}
\begin{subfigure}[t]{0.15\textwidth}
\includegraphics[width=\linewidth]{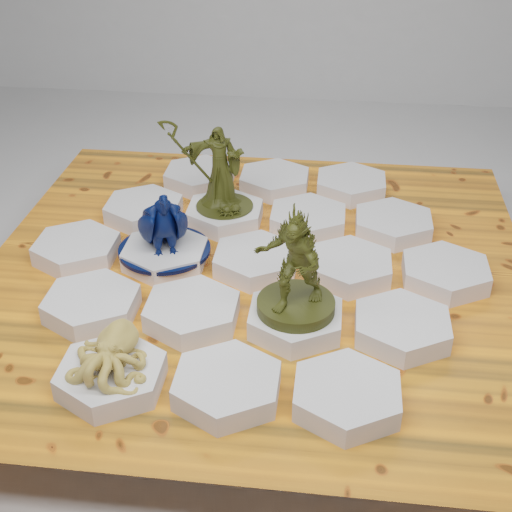}
\end{subfigure}
\begin{subfigure}[t]{0.15\textwidth}
\includegraphics[width=\linewidth]{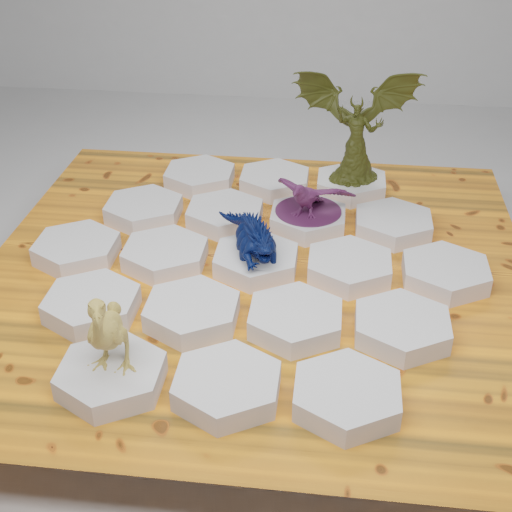}
\end{subfigure}
\begin{subfigure}[t]{0.15\textwidth}
\includegraphics[width=\linewidth]{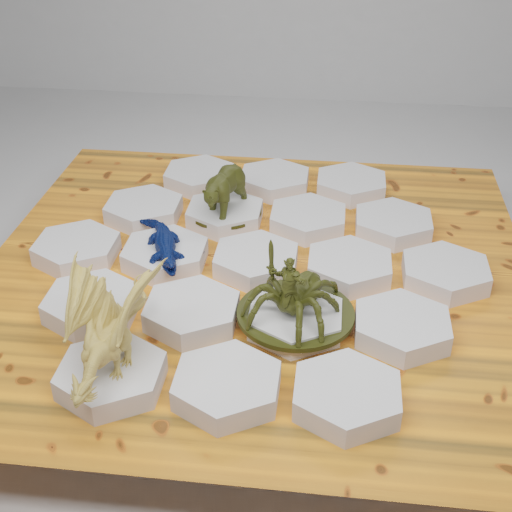}
\end{subfigure}
\begin{subfigure}[t]{0.15\textwidth}
\includegraphics[width=\linewidth]{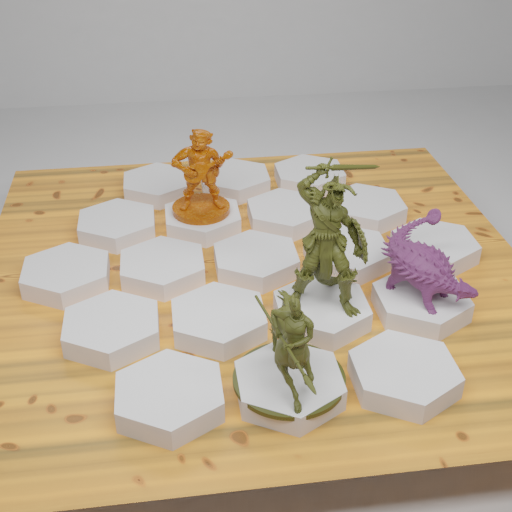}
\end{subfigure}
\begin{subfigure}[t]{0.15\textwidth}
\includegraphics[width=\linewidth]{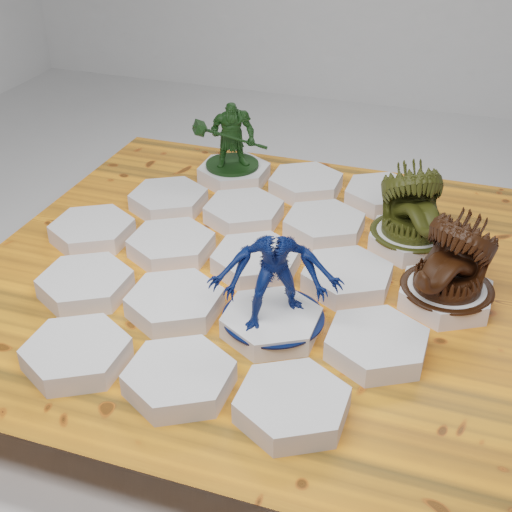}
\end{subfigure}
\begin{subfigure}[t]{0.15\textwidth}
\includegraphics[width=\linewidth]{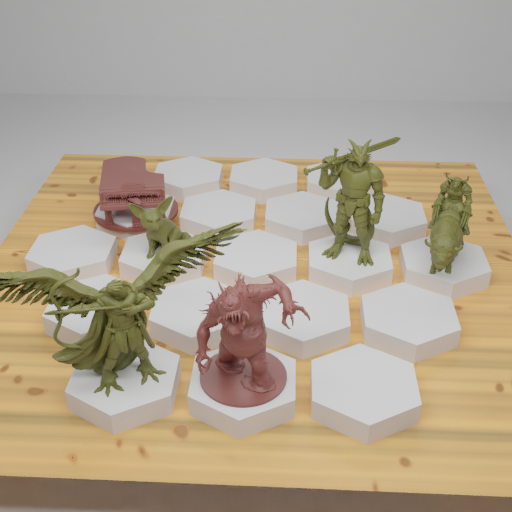}
\end{subfigure}
\begin{subfigure}[t]{0.15\textwidth}
\includegraphics[width=\linewidth]{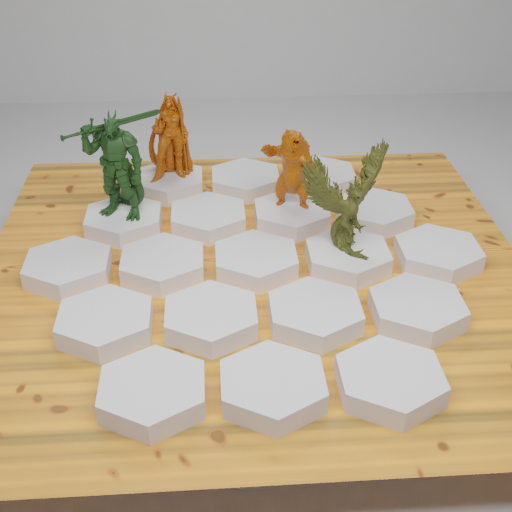}
\end{subfigure}
\begin{subfigure}[t]{0.15\textwidth}
\includegraphics[width=\linewidth]{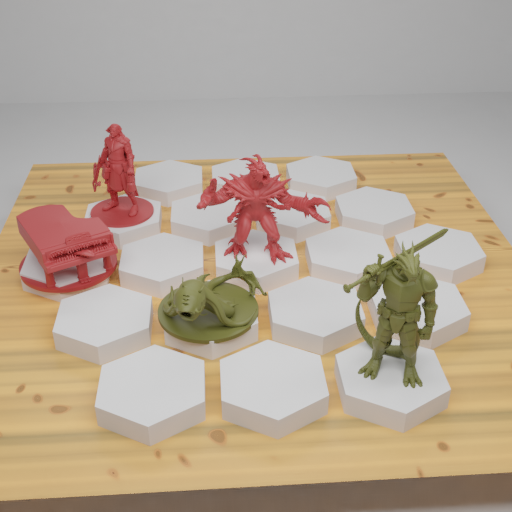}
\end{subfigure}
\begin{subfigure}[t]{0.15\textwidth}
\includegraphics[width=\linewidth]{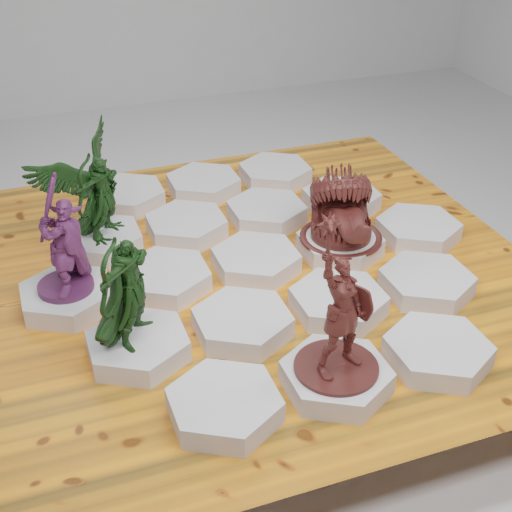}
\end{subfigure}
\begin{subfigure}[t]{0.15\textwidth}
\includegraphics[width=\linewidth]{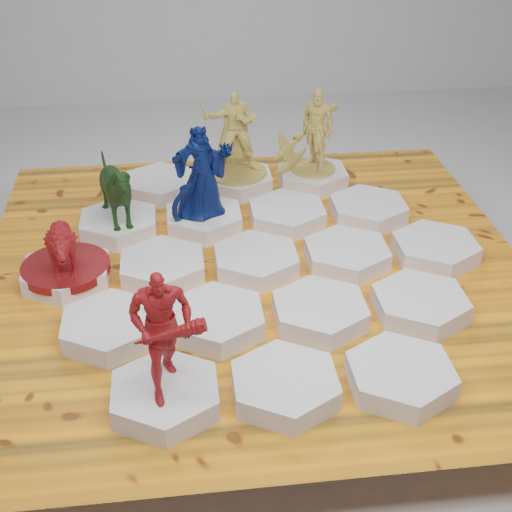}
\end{subfigure}
\begin{subfigure}[t]{0.15\textwidth}
\includegraphics[width=\linewidth]{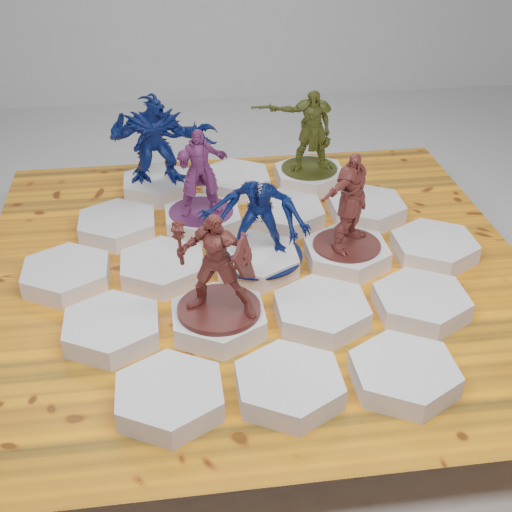}
\end{subfigure}
\begin{subfigure}[t]{0.15\textwidth}
\includegraphics[width=\linewidth]{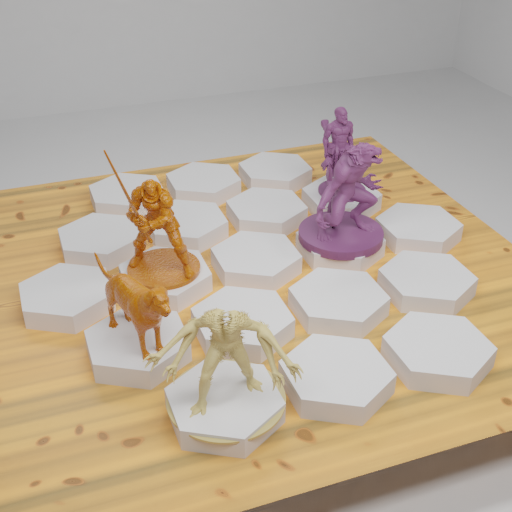}
\end{subfigure}
\begin{subfigure}[t]{0.15\textwidth}
\includegraphics[width=\linewidth]{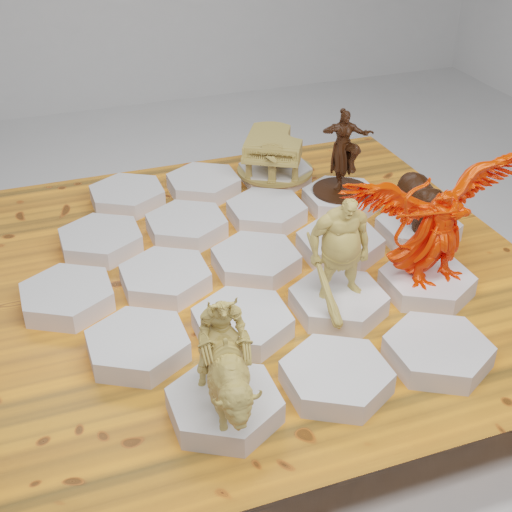}
\end{subfigure}
\begin{subfigure}[t]{0.15\textwidth}
\includegraphics[width=\linewidth]{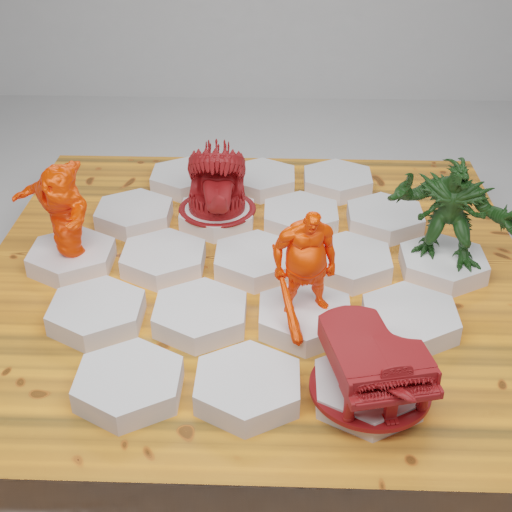}
\end{subfigure}
\begin{subfigure}[t]{0.15\textwidth}
\includegraphics[width=\linewidth]{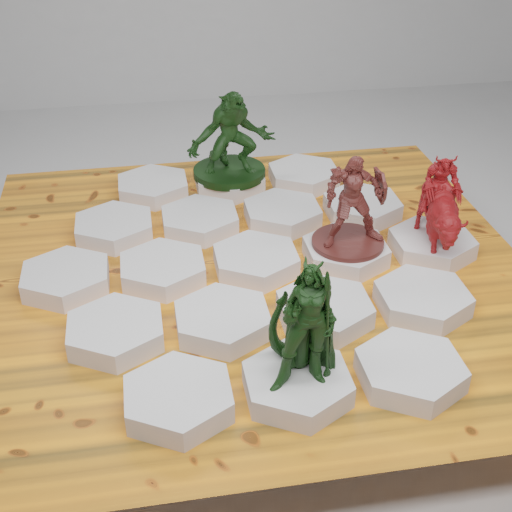}
\end{subfigure}
\begin{subfigure}[t]{0.15\textwidth}
\includegraphics[width=\linewidth]{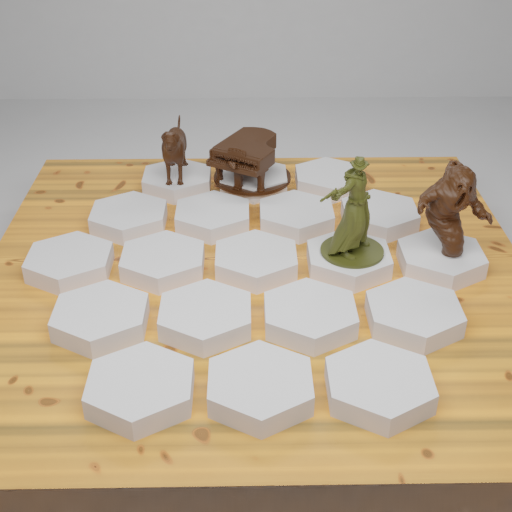}
\end{subfigure}
\begin{subfigure}[t]{0.15\textwidth}
\includegraphics[width=\linewidth]{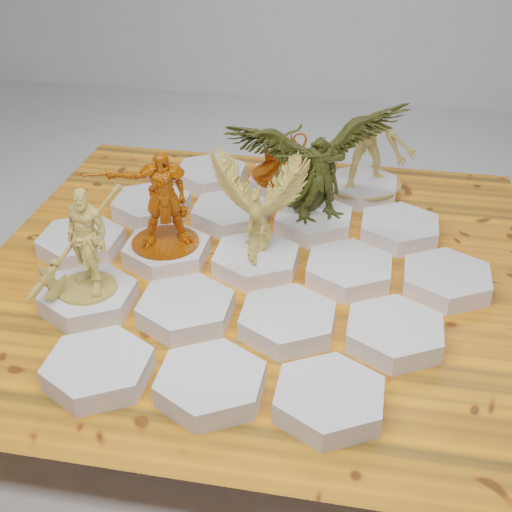}
\end{subfigure}
\begin{subfigure}[t]{0.15\textwidth}
\includegraphics[width=\linewidth]{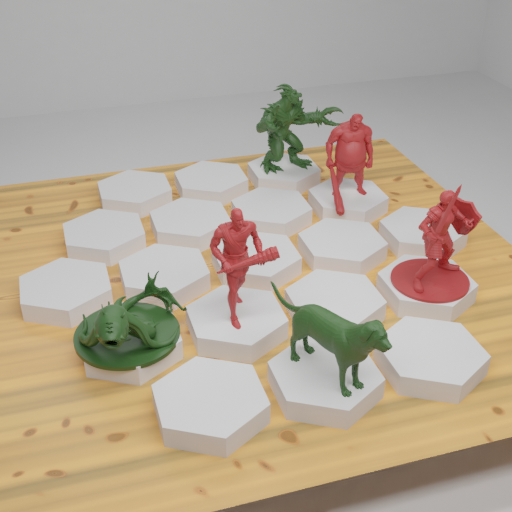}
\end{subfigure}
\begin{subfigure}[t]{0.15\textwidth}
\includegraphics[width=\linewidth]{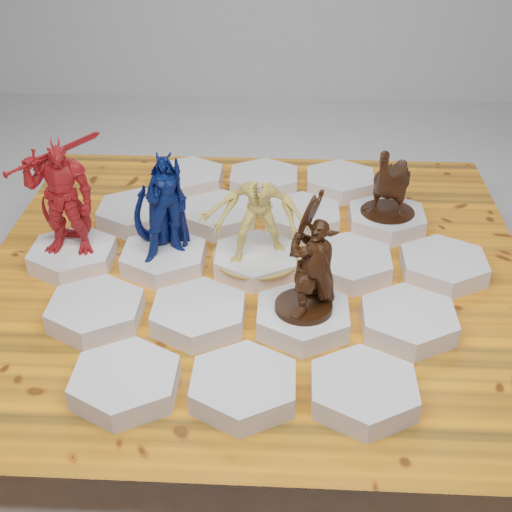}
\end{subfigure}
\begin{subfigure}[t]{0.15\textwidth}
\includegraphics[width=\linewidth]{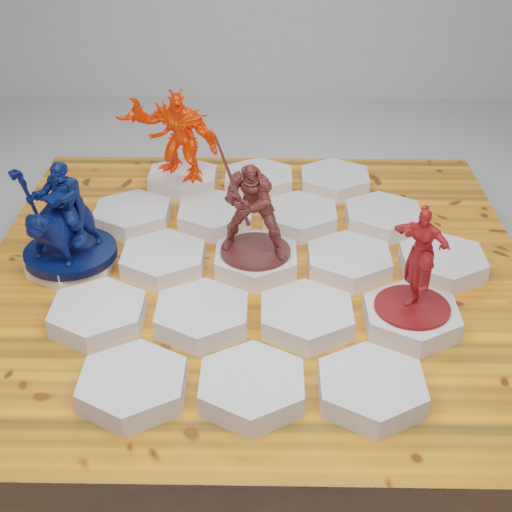}
\end{subfigure}
\begin{subfigure}[t]{0.15\textwidth}
\includegraphics[width=\linewidth]{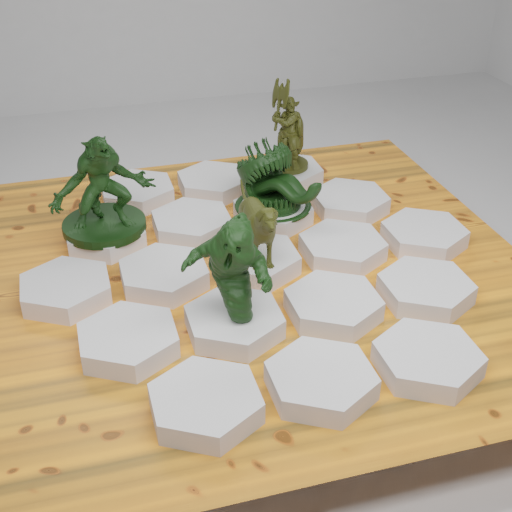}
\end{subfigure}
\begin{subfigure}[t]{0.15\textwidth}
\includegraphics[width=\linewidth]{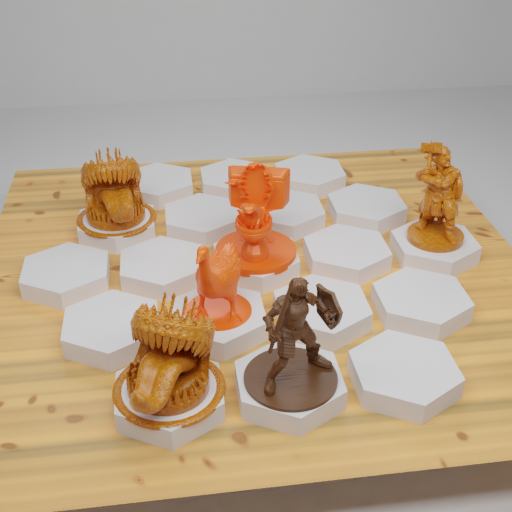}
\end{subfigure}
\begin{subfigure}[t]{0.15\textwidth}
\includegraphics[width=\linewidth]{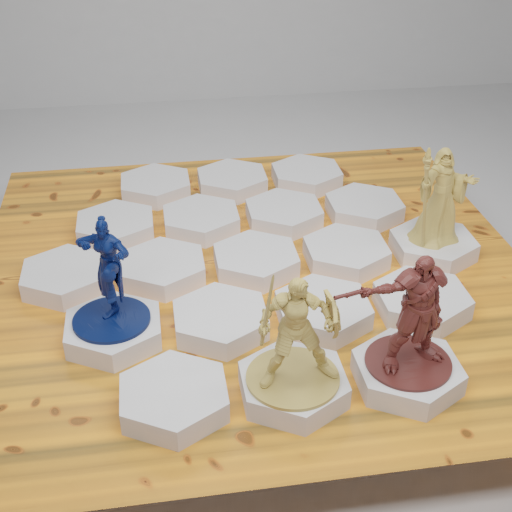}
\end{subfigure}
\begin{subfigure}[t]{0.15\textwidth}
\includegraphics[width=\linewidth]{images/group1scene1001.png}
\end{subfigure}
\begin{subfigure}[t]{0.15\textwidth}
\includegraphics[width=\linewidth]{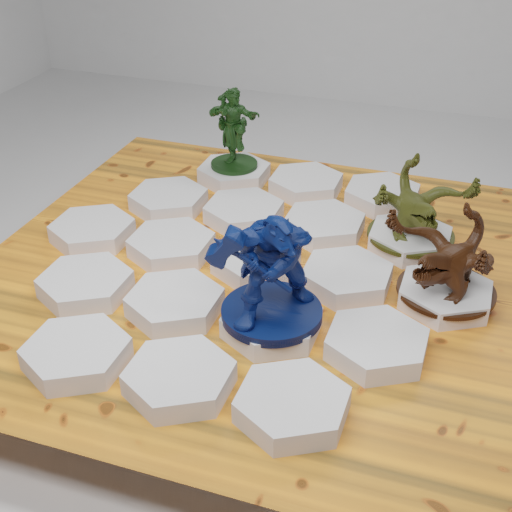}
\end{subfigure}
\begin{subfigure}[t]{0.15\textwidth}
\includegraphics[width=\linewidth]{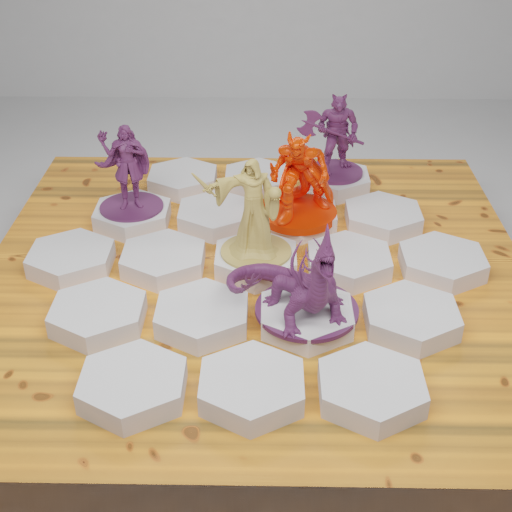}
\end{subfigure}
\begin{subfigure}[t]{0.15\textwidth}
\includegraphics[width=\linewidth]{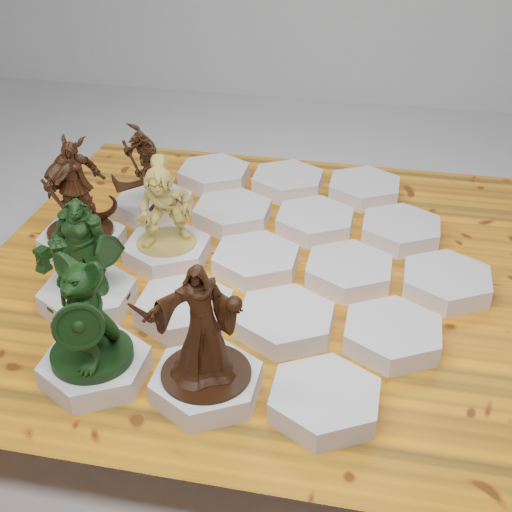}
\end{subfigure}
\begin{subfigure}[t]{0.15\textwidth}
\includegraphics[width=\linewidth]{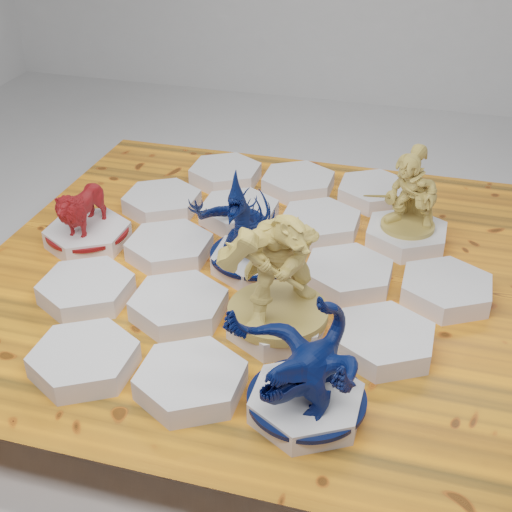}
\end{subfigure}
\begin{subfigure}[t]{0.15\textwidth}
\includegraphics[width=\linewidth]{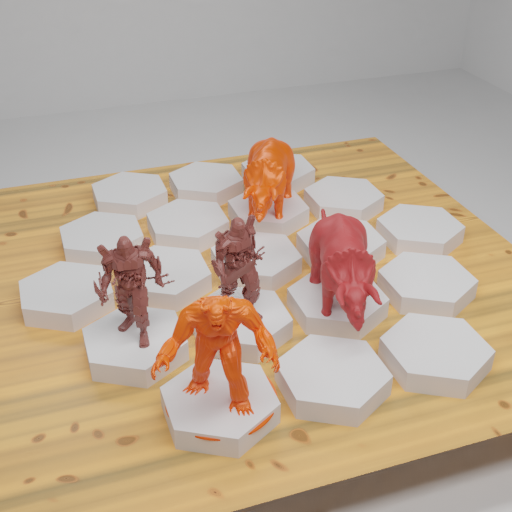}
\end{subfigure}
\begin{subfigure}[t]{0.15\textwidth}
\includegraphics[width=\linewidth]{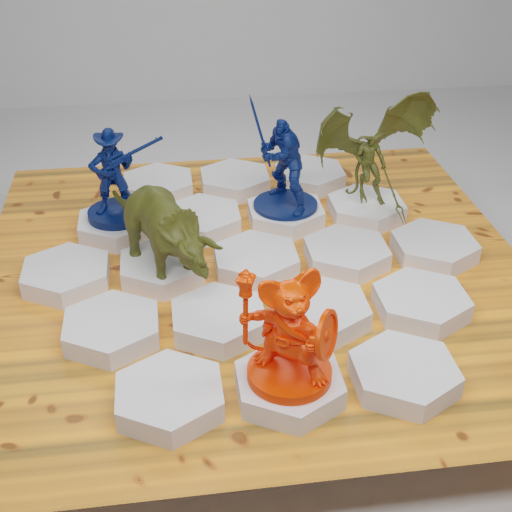}
\end{subfigure}
\begin{subfigure}[t]{0.15\textwidth}
\includegraphics[width=\linewidth]{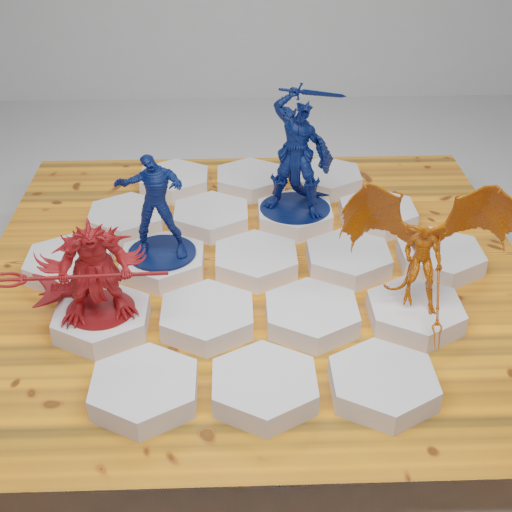}
\end{subfigure}
\begin{subfigure}[t]{0.15\textwidth}
\includegraphics[width=\linewidth]{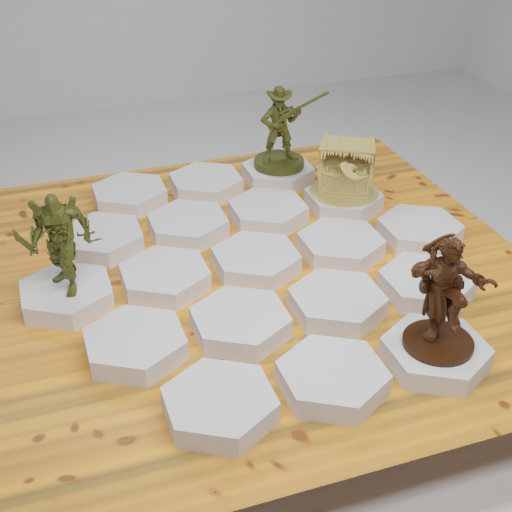}
\end{subfigure}
\begin{subfigure}[t]{0.15\textwidth}
\includegraphics[width=\linewidth]{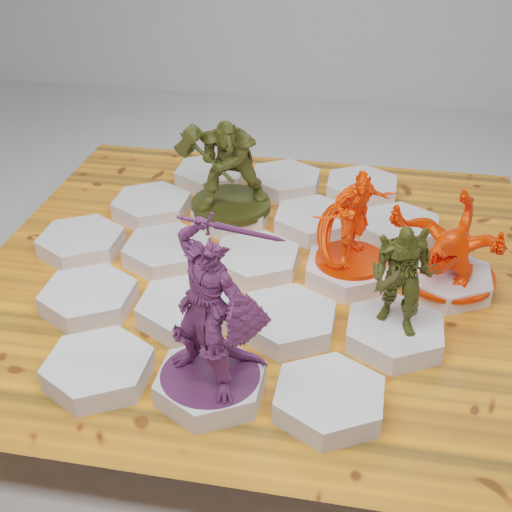}
\end{subfigure}
\begin{subfigure}[t]{0.15\textwidth}
\includegraphics[width=\linewidth]{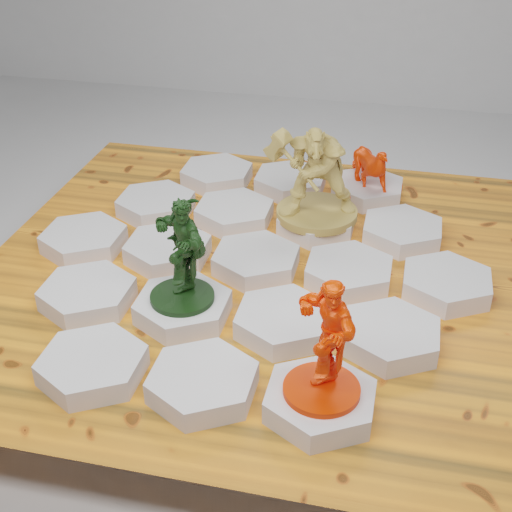}
\end{subfigure}
\begin{subfigure}[t]{0.15\textwidth}
\includegraphics[width=\linewidth]{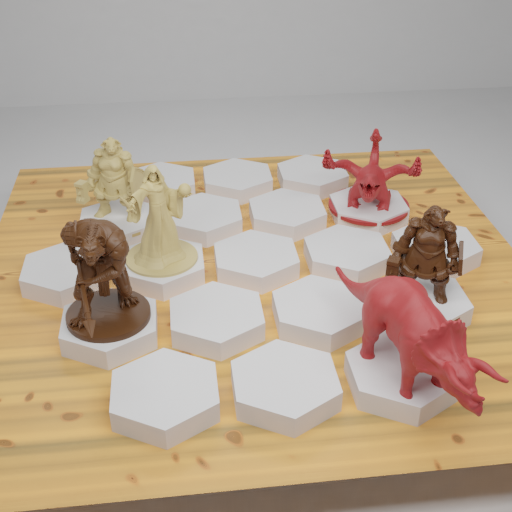}
\end{subfigure}
\begin{subfigure}[t]{0.15\textwidth}
\includegraphics[width=\linewidth]{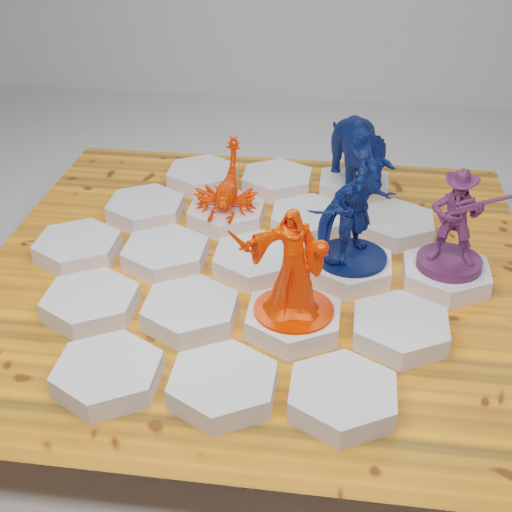}
\end{subfigure}
\begin{subfigure}[t]{0.15\textwidth}
\includegraphics[width=\linewidth]{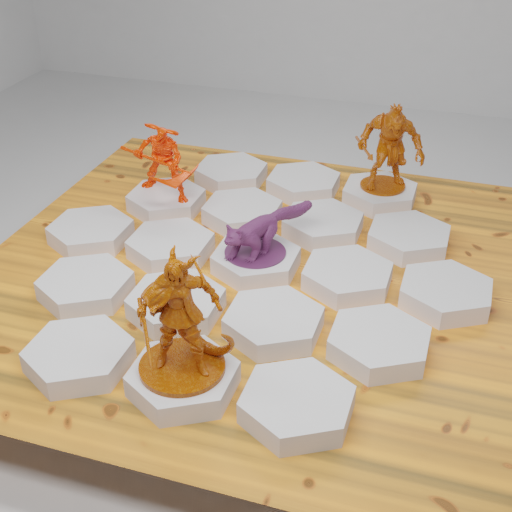}
\end{subfigure}
\begin{subfigure}[t]{0.15\textwidth}
\includegraphics[width=\linewidth]{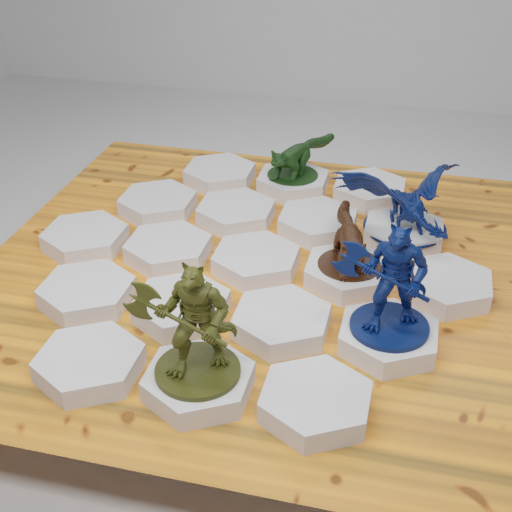}
\end{subfigure}
\begin{subfigure}[t]{0.15\textwidth}
\includegraphics[width=\linewidth]{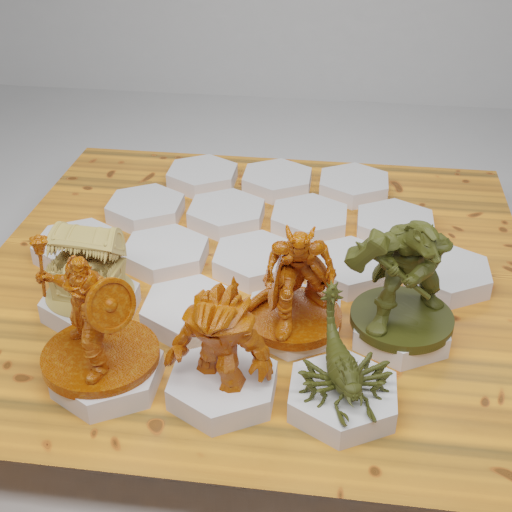}
\end{subfigure}
\begin{subfigure}[t]{0.15\textwidth}
\includegraphics[width=\linewidth]{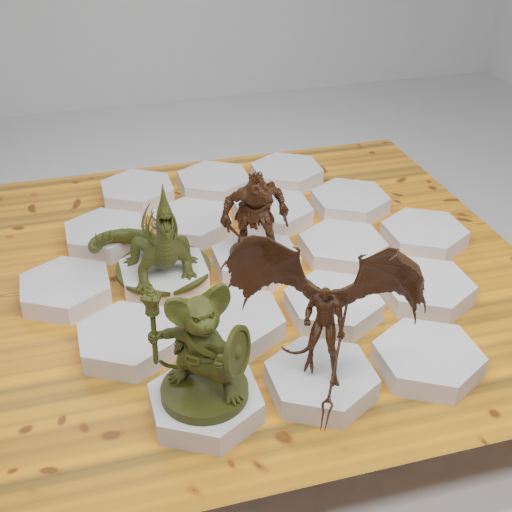}
\end{subfigure}
\begin{subfigure}[t]{0.15\textwidth}
\includegraphics[width=\linewidth]{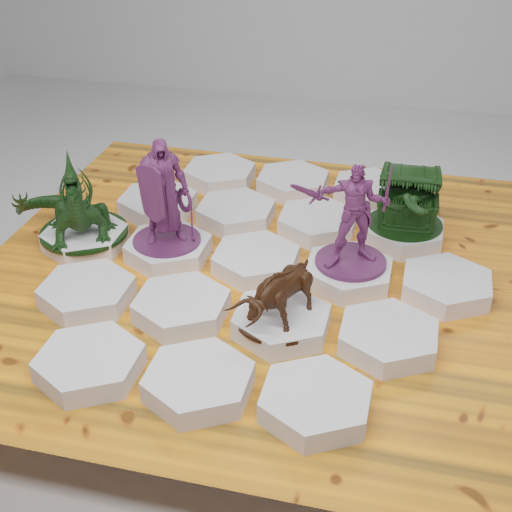}
\end{subfigure}
\begin{subfigure}[t]{0.15\textwidth}
\includegraphics[width=\linewidth]{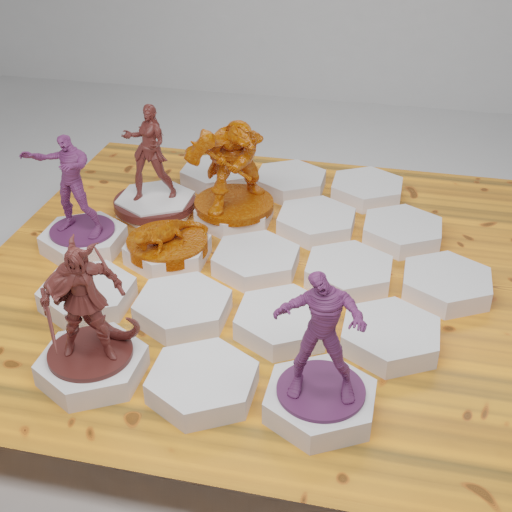}
\end{subfigure}
\begin{subfigure}[t]{0.15\textwidth}
\includegraphics[width=\linewidth]{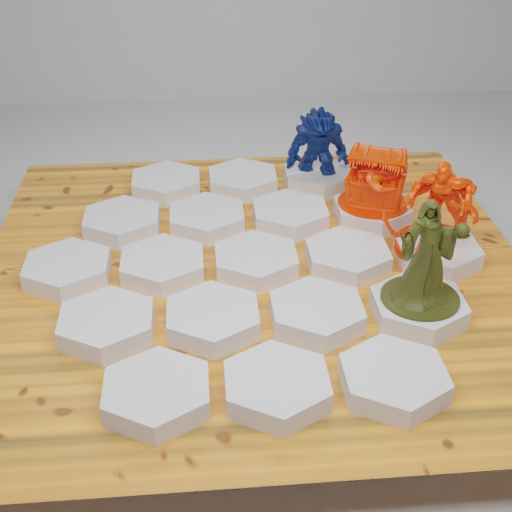}
\end{subfigure}
\begin{subfigure}[t]{0.15\textwidth}
\includegraphics[width=\linewidth]{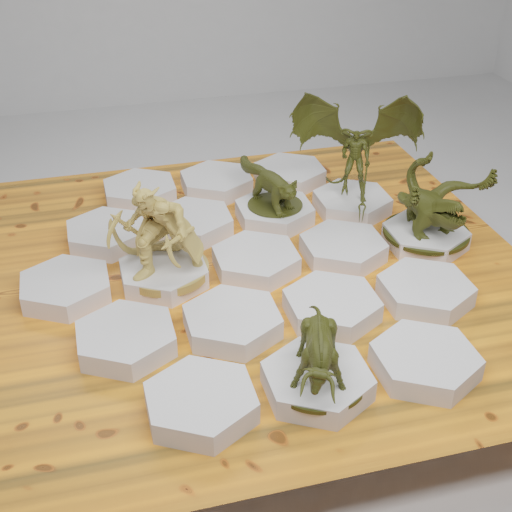}
\end{subfigure}
\caption{Examples scenes of UOUC.}
\label{image}
\end{center}
\end{figure*}

\begin{figure}
    \centering
    \begin{subfigure}[t]{0.15\textwidth}
    \includegraphics[width=\linewidth]{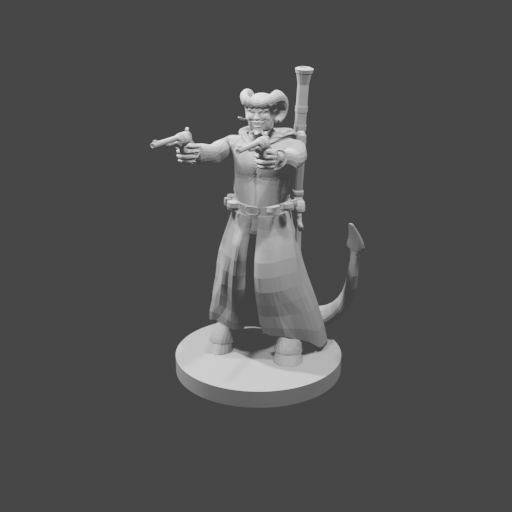}
    \caption{}
    \end{subfigure}
    \begin{subfigure}[t]{0.15\textwidth}
    \includegraphics[width=\linewidth]{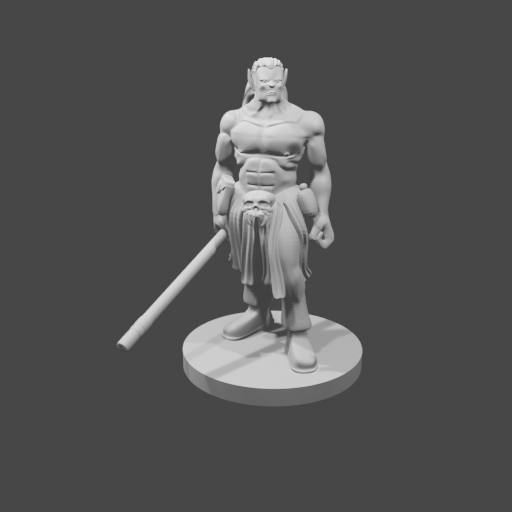}
    \caption{}
    \end{subfigure}
    \begin{subfigure}[t]{0.15\textwidth}
    \includegraphics[width=\linewidth]{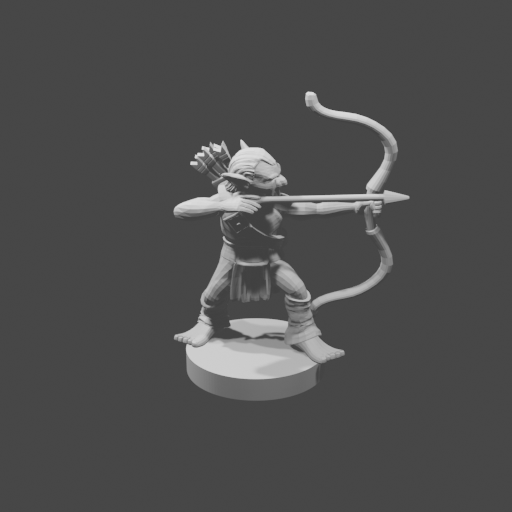}
    \caption{}
    \end{subfigure}
    \begin{subfigure}[t]{0.15\textwidth}
    \includegraphics[width=\linewidth]{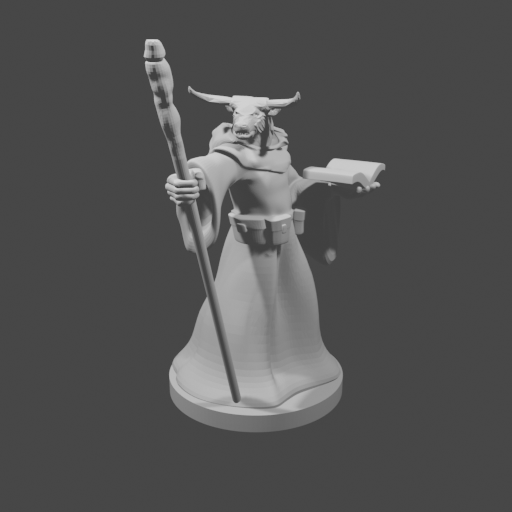}
    \caption{}
    \end{subfigure}
    \begin{subfigure}[t]{0.15\textwidth}
    \includegraphics[width=\linewidth]{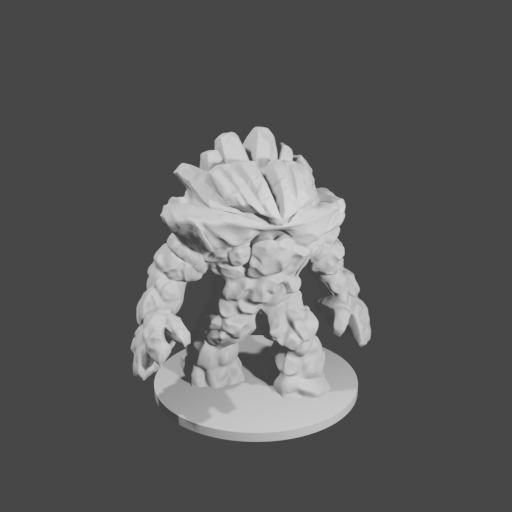}
    \caption{}
    \end{subfigure}
    \begin{subfigure}[t]{0.15\textwidth}
    \includegraphics[width=\linewidth]{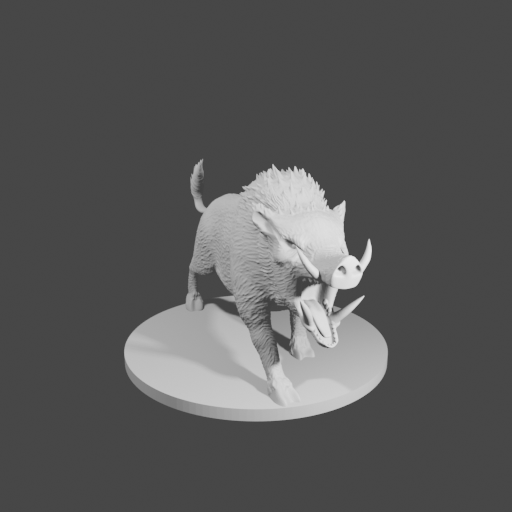}
    \caption{}
    \end{subfigure}
    \begin{subfigure}[t]{0.15\textwidth}
    \includegraphics[width=\linewidth]{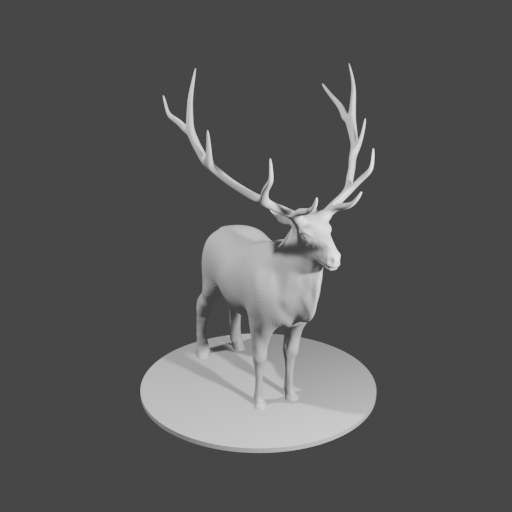}
    \caption{}
    \end{subfigure}
    \caption{Examples of objects in UOUC. Note the complexity and diversity of appearance.}
    \label{fig:object}
\end{figure}

We present more examples of scenes and objects, so that the complexity of UOUC may be appreciated better. Figure \ref{image} presents examples of scenes. Please note that while the base scene is not complex, the individual objects exhibit great complexity. Figure \ref{fig:object} presents examples of objects at a better resolution for a better feel of the appearance.

\section{Some further statistics for UOUC (in reference to section 4)}
We present statistics relating to the extent of uniformity in the presence of attributes in the scenes of the train and test set, and also in the answers of each question-type.

\begin{figure*}
    \centering
    \begin{subfigure}[t]{\textwidth}
    \includegraphics[width=\linewidth]{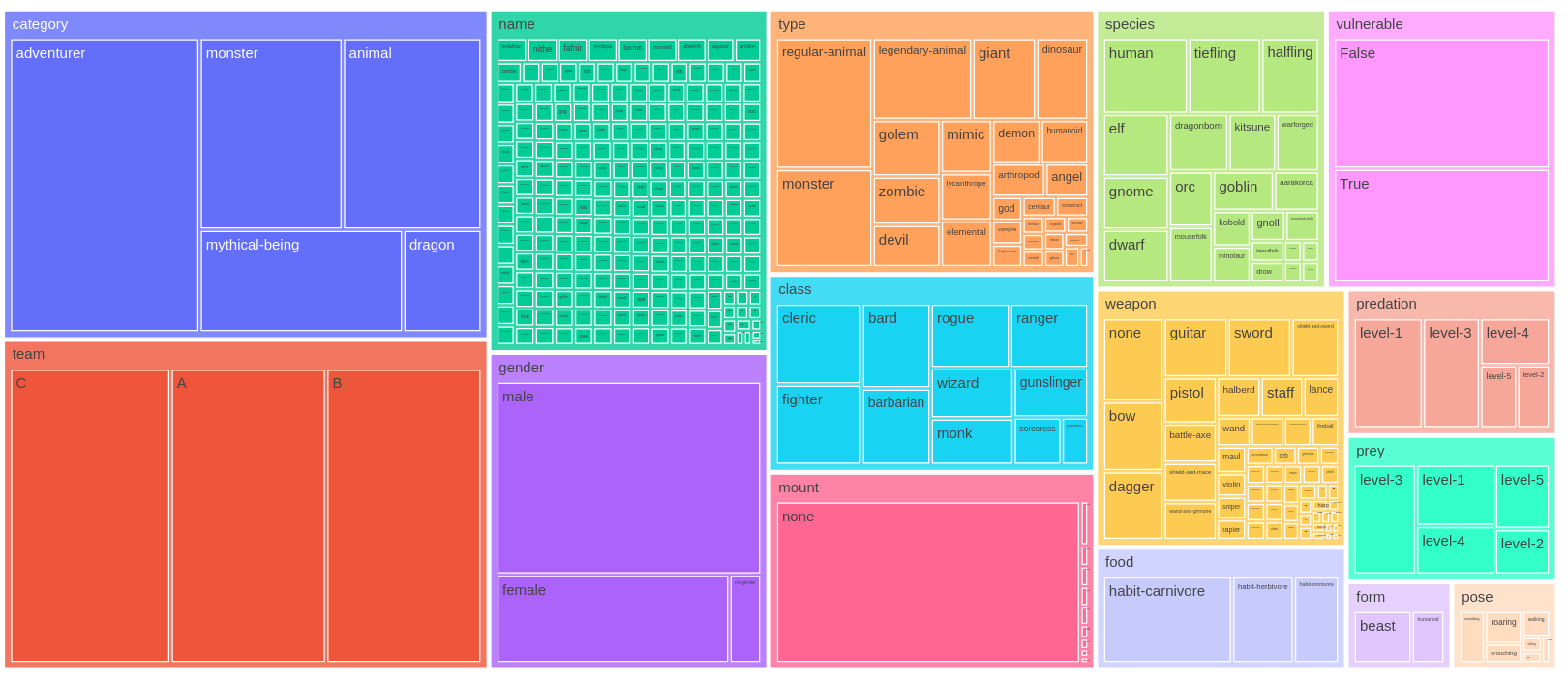}
    \caption{The distribution of properties among objects of the train data.}
    \label{fig:my_label1}
    \end{subfigure}
    \begin{subfigure}[t]{\textwidth}
    \includegraphics[width=\linewidth]{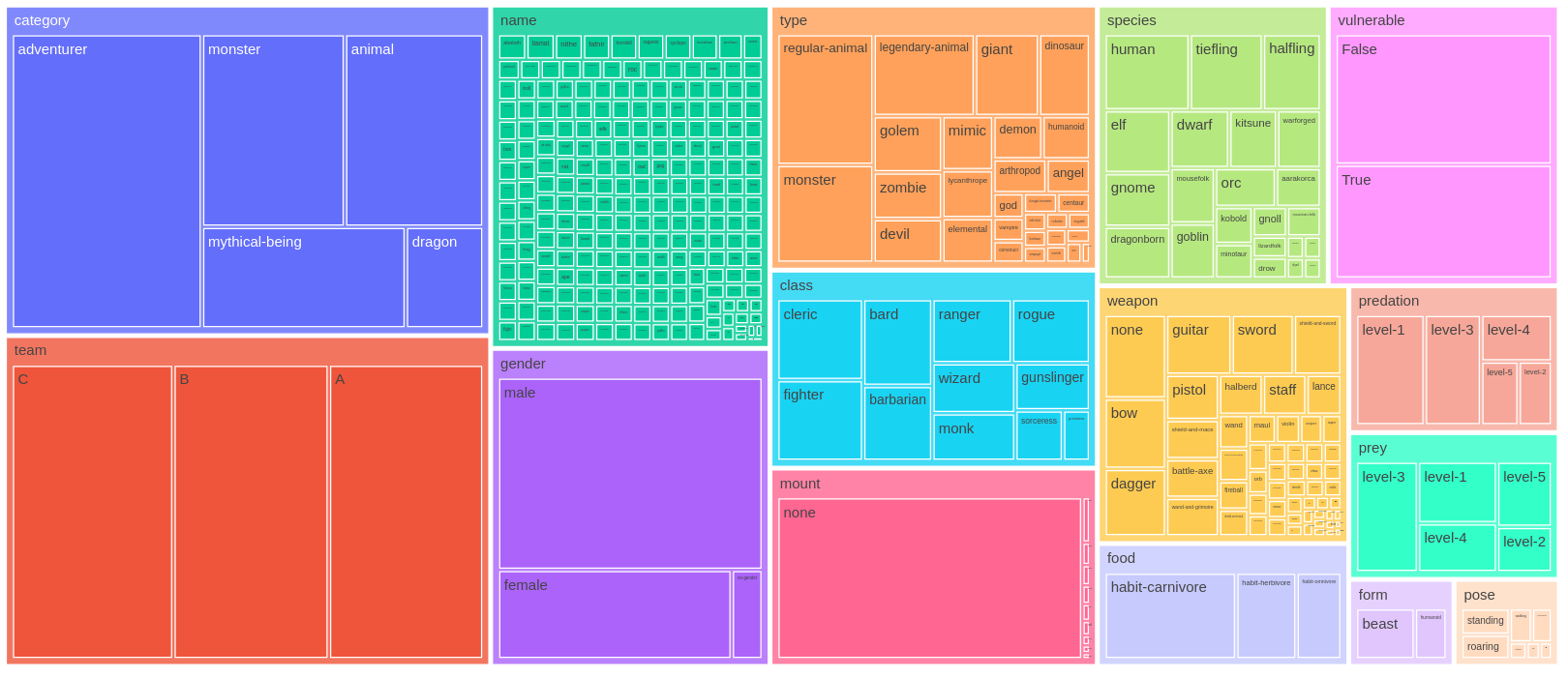}
    \caption{The distribution of properties among objects of the test data.}
    \label{fig:my_label2}
    \end{subfigure}
    \caption{Larger rectangles indicate a greater proportion of instances of the properties (figures best viewed in colour).}
\end{figure*}

In \ref{fig:my_label1}, we provide a representation of the relative numbers of objects with certain attributes for the train data. In \ref{fig:my_label2}, a representation of such statistics is given for the test data set. Each rectangle of a colour, within the figures, is representative of an attribute. Boxes within the rectangles represent specific values taken by the corresponding attribute. For example, consider the blue rectangle for the train data. This rectangle represents the category of the objects in the scene. The size of the boxes in this rectangle represent the proportions of the occurrences of the values of the categories. A larger size of a box indicates a larger proportion. A well-balanced attribute would have boxes of similar sizes. While some boxes are labelled, others which relate to attributes with lower proportions are not labelled.

We see that most objects, in the train and test data, are categorised as adventurers. This is because most objects are adventurers. However, objects from other categories also are present in a noticeable fraction. Thus, while adventurer-objects are present in most scenes, objects of other categories are likely to be present in most scenes as well.

We see that most objects that have a gender-property are given 'male' as the property. This was not by conscious design. Most objects of categories that were associated with a gender had features generally accepted to be of male gender. We believe that this data set has no direct application in scenarios where gender-bias could affect people, and hence believe it has no negative social impact. This much said, we would definitely aim at an extension of the object-class of UOUC to have it gender-balanced.

We see that there is a significant imbalance in the objects having mounts - most do not. Observing the presence of certain number of attributes that occur infrequently in objects, it can be expected that even for memorisation-questions, for high performance, models must be able to have a means of keeping such information. A test for this would be to construct a test data set, consisting of objects with these properties. This has not been done in our work, as compositionality is our main area of study. The fact that the train and test sets have a similar distribution of attributes, indicates that no additional difficulty has been introduced beyond object-co-occurrence.

\begin{figure*}
\centering
\begin{subfigure}[t]{\textwidth}
\includegraphics[width=\linewidth]{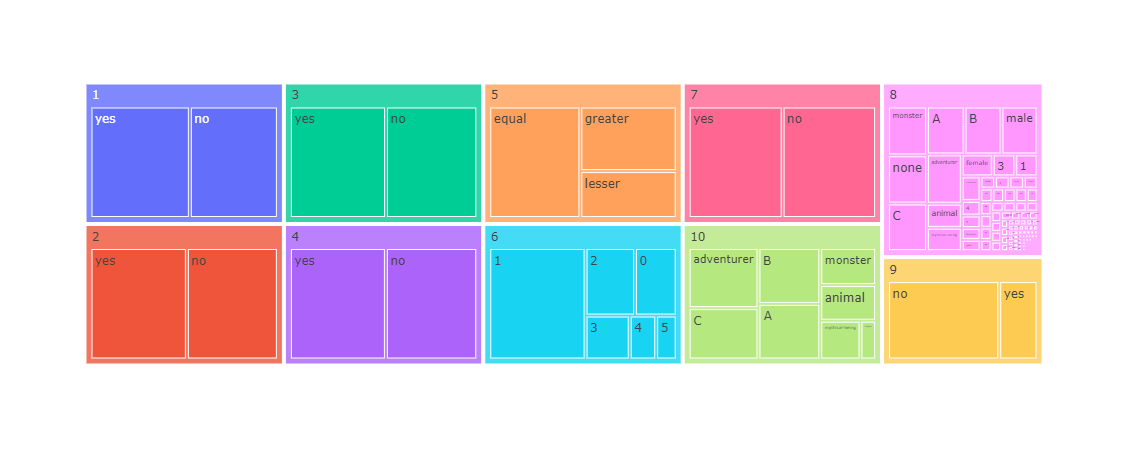}
\caption{The distribution of answers among questions of the train data}
\label{fig:label}
\end{subfigure}
\begin{subfigure}[t]{\textwidth}
\includegraphics[width=\linewidth]{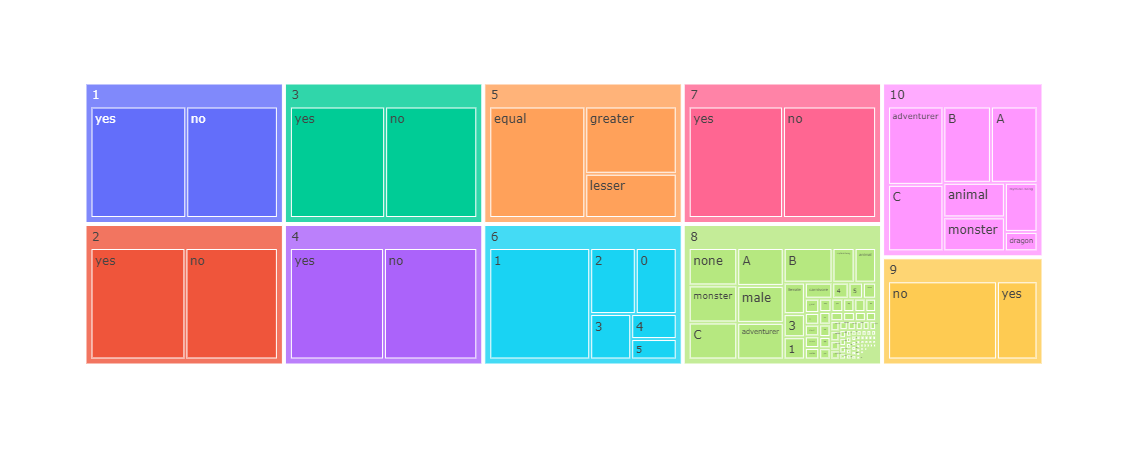}
\caption{The distribution of properties among questions of the test data}
\label{fig:label1}
\end{subfigure}
\caption{Larger rectangles indicate a greater proportion of answers of a question-type (figures best viewed in colour).}
\end{figure*}

In \ref{fig:label}, we present a representation of the uniformity of answers for each question-type for the train scenes. In \ref{fig:label1}, we present the statistics for the test scenes. The representation is similarly interpreted as in the representation of attributes. First, we see that question 1, 2, 3, 4, and 7 are all well-balanced. Each of these has only two possible answers. Question 9, which deals with non-spatial, memory-based relationships, is not as balanced in its answer-distribution as most pairs of animal-animal and animal-adventurer are non-attacking. This comes from a natural consequence of having fewer animals with a large predation-level.

Question 5 is not particularly poor in its balance, even if one of its answers occurs more frequently than the others. This is because its other answers also occur at a good proportion. Question 10 and Question 6, on the other hand, have certain answers that occur, comparatively, infrequently.

Question 8 has multiple possible answers, many of which occur infrequently. This is related to the fact that the objects themselves have a similar distribution of attributes. Again, we see that the train and test sets are similar in distribution, indicating no added difficulty in the test set apart from the co-occurrence of objects.

\section{Some other details about the training of models (in reference to section 5)}
The following are the sources from where we took the implementations of the models we used for VQA. We used the pytorch implementation of MAC from  \url{https://github.com/rosinality/mac-network-pytorch}, SAAA from \cite{kazemi2017show} from \url{https://github.com/Cyanogenoid/pytorch-vqa}, the pytorch implementation of MUTAN from \url{https://github.com/Cadene/vqa.pytorch}, and LCGN from \url{https://github.com/ronghanghu/lcgn/tree/pytorch}.

We trained MAC, MUTAN, LCGN, and SAAA for 50 epochs. We trained each of the image-only and language-only models for 100 epochs. Other hyperparameters such as the learning rate, the number of layers, the scheduling of lr etc. are similar to what is used in the original source.

\bibliographystyle{plain}
\bibliography{egbib}
\end{document}